\documentclass[mlmain]{jmlr}

\usepackage{longtable}

\usepackage{booktabs}
\usepackage[load-configurations=version-1]{siunitx} 

\theorembodyfont{\upshape}
\theoremheaderfont{\scshape}
\theorempostheader{:}
\theoremsep{\newline}

\jmlrvolume{}
\firstpageno{1}
\editors{List of editors' names}

\jmlryear{2024}
\jmlrworkshop{Symmetry and Geometry in Neural Representations}

\allowdisplaybreaks

\usepackage{amsmath,amsfonts,bm}
\usepackage{xcolor}
\usepackage{fontawesome}

\def\eqref#1{(\ref{#1})}

\def\1{\bm{1}}

\def\rvb{{\mathbf{b}}}

\def\rve{{\mathbf{e}}}

\def\rvu{{\mathbf{i}}}

\def\rvo{{\mathbf{o}}}

\def\rvu{{\mathbf{u}}}
\def\rvv{{\mathbf{v}}}
\def\rvw{{\mathbf{w}}}
\def\rvx{{\mathbf{x}}}
\def\rvy{{\mathbf{y}}}
\def\rvz{{\mathbf{z}}}

\def\rmI{{\mathbf{I}}}

\def\rmM{{\mathbf{M}}}

\def\rmW{{\mathbf{W}}}

\def\vzero{{\bm{0}}}
\def\vone{{\bm{1}}}

\DeclareMathAlphabet{\mathsfit}{\encodingdefault}{\sfdefault}{m}{sl}
\SetMathAlphabet{\mathsfit}{bold}{\encodingdefault}{\sfdefault}{bx}{n}

\def\gB{{\mathcal{B}}}

\def\gK{{\mathcal{K}}}
\def\gL{{\mathcal{L}}}
\def\gM{{\mathcal{M}}}

\def\gO{{\mathcal{O}}}

\def\gT{{\mathcal{T}}}

\def\sB{{\mathbb{B}}}

\def\sH{{\mathbb{H}}}

\def\sK{{\mathbb{K}}}
\def\sL{{\mathbb{L}}}

\def\sR{{\mathbb{R}}}

\def\sV{{\mathbb{V}}}

\newcommand{\R}{\mathbb{R}}

\newcommand{\norm}[1]{\left\lVert#1\right\rVert}
\newcommand{\T}{\top}

\newcommand{\qed}{\hfill \ensuremath{\blacksquare}}

\usepackage[textsize=tiny, backgroundcolor=blue!30]{todonotes}
\usepackage{listings}
\usepackage{color}

\usepackage{comment}

\title[Klein HNN]{Klein Model for Hyperbolic Neural Networks 
}

 \author{\Name{Yidan Mao} \Email{ym188@duke.edu}\\
 \addr Duke Kunshan University
 \AND
 \Name{Jing Gu} \Email{gu003981@umn.edu}\\
 \addr University of Minnesota, Twin Cities
 \AND
 \Name{Marcus C. Werner} \Email{marcus.werner@duke.edu}\\
 \addr Duke Kunshan University
 \AND
 \Name{Dongmian Zou} \Email{dongmian.zou@duke.edu}\\
 \addr Duke Kunshan University
 }

\begin{document}

\maketitle

\begin{abstract}
Hyperbolic neural networks (HNNs) have been proved effective in modeling complex data structures. However, previous works mainly focused on the Poincar\'e ball model and the hyperboloid model as coordinate representations of the hyperbolic space, often neglecting the Klein model. Despite this, the Klein model offers its distinct advantages thanks to its straight-line geodesics, which facilitates the well-known Einstein midpoint construction, previously leveraged to accompany HNNs in other models. In this work, we introduce a framework for hyperbolic neural networks based on the Klein model. We provide detailed formulation for representing useful operations using the Klein model. We further study the Klein linear layer and prove that the ``tangent space construction'' of the scalar multiplication and parallel transport are exactly the Einstein scalar multiplication and the Einstein addition, analogous to the M\"obius operations used in the Poincar\'e ball model. We show numerically that the Klein HNN performs on par with the Poincar\'e ball model, providing a third option for HNN that works as a building block for more complicated architectures. 
\end{abstract}
\begin{keywords}
Hyperbolic neural network, Klein model, Einstein gyrovector space 
\end{keywords}

\section{Introduction}\label{sec:intro}
Hyperbolic spaces have shown considerable promise in embedding complex networks~\citep{krioukov2010hyperbolic}, trees~\citep{wilson2014spherical, sonthalia2020tree} and hierarchical datasets~\citep{nickel2018learning}. To leverage the inherent geometric structures within these data types in learning neural representations, hyperbolic neural networks (HNNs) were introduced, initially by \citet{ganea2018hyperbolic} and later expanded upon by many recent works~\citep{peng2021hyperbolic}. To define neural operations in hyperbolic spaces, HNNs typically utilize a model of hyperbolic geometry, where points are represented by Euclidean coordinates. Most of the literature focuses on either the Poincar\'e ball model or the hyperboloid model. Indeed, \citet{ganea2018hyperbolic, shimizu2021hyperbolic} presented a set of operations for implementing HNNs using the Poincar\'e ball model, which often enjoys a simple mathematical description. On the other hand, the hyperboloid model proves to be numerically more stable than the Poincar\'e ball model~\citep{nickel2018learning, qu2022hyperbolic, mishne2023numerical} and facilitates hyperbolic ``linear'' layers without using tangent spaces~\citep{chen2021fully}.

Despite the success of HNNs using the Poincar\'e ball and the hyperboloid models, HNNs using other models remain largely unexplored. This raises problems when important operations are defined only in a certain model. Indeed, in many works \citep{gulcehre2018hyperbolic, zhu2020hypertext, khrulkov2020hyperbolic, zhang2021hype, tai2021knowledge, song2022hyperbolic, fu2024hyperbolic, skrodzki2024accelerating, li2024dhgat}, in order to use the Einstein midpoint to perform aggregation, a mapping between the Poincar\'e ball/hyperboloid model and the Klein model has to be implemented whenever aggregation is needed. This causes unnecessary computational complexity. The Klein model also enjoys other crucial properties such as straight-line geodesics. However, to the best of our knowledge, no prior works have discussed compact neural operations using the Klein model, that is, ``pure Klein'' HNNs. Moreover, recently developed software packages for HNNs lack implementation in the Klein model~\citep{van2023hypll}.

In this paper, we derive compact formulas for key operations in the Klein model. Analogous to the M\"obius operations in the Poincar\'e ball model, we show that the Klein model facilitates a set of operations known as Einstein scalar multiplication and Einstein addition~\citep{ungar2022gyrovector, ungar2012beyond}. For the Poincar\'e ball model, \citet{ganea2018hyperbolic} demonstrated that the M\"obius operations are exactly the exponential maps of ``linear'' operations (as used in multilayer perceptrons for obtaining pre-activations) in tangent spaces. However, what remains unclear is the precise relationship between Einstein operations and similar ``tangent space constructions'' in the Klein model. In our paper, we prove that they are equivalent operations. Thanks to this interpretation, we build Klein HNNs based on these Einstein operations. Through experiments on well-known graph datasets, we show that the performance of Klein HNNs is on par with both Poincar\'e HNNs and hyperboloid HNNs. Furthermore, Klein HNNs are efficient to implement.

\section{Related Works}\label{sec:related_works}
\paragraph{Hyperbolic neural networks} 
The first HNN was introduced relatively recently by \citet{ganea2018hyperbolic}, who utilized the Poincar\'e ball model to develop fundamental neural operations including linear layers and recurrent layers. Importantly, they demonstrated that the M\"obius operations correspond to the tangent space operations that one would intuitively define for neural operations in the Poincar\'e ball model. Later, \citet{shimizu2021hyperbolic} extended upon their work and introduced more operations. Another line of research employed the hyperboloid model, favored for its numerical stability~\citep{gulcehre2018hyperbolic, chen2021fully}. More complex hyperbolic neural operations have since been designed for a variety of applications, particularly for graph data~\citep{chami2019hyperbolic, liu2019hyperbolic, mathieu2019continuous, dai2021hyperbolic, zhang2021lorentzian, yang2022hyperbolic, yang2024hypformer} and image data~\citep{khrulkov2020hyperbolic, atigh2022hyperbolic, ermolov2022hyperbolic, desai2023hyperbolic, yang2024improving}. Additionally, there have been studies that focused on the numerical stability~\citep{mishne2023numerical} and robustness~\citep{li2024improving} of HNNs. Despite these advances, most models continue to rely on either the Poincaré ball or hyperboloid models, with other hyperbolic models largely overlooked.

\paragraph{The Klein model in use} 
Recently, the Klein model has been proved effective in representing various types of data, including hierarchical graphs~\citep{mcdonald2020heat, yang2024towards}, protein sequences~\citep{ali2024gaussian}, minimal spanning trees~\citep{garcia2024hypersteiner}, and scene images~\citep{bi2017multiple}. 
A key advantage of the Klein model is that its geodesics are represented as straight lines, which facilitates intuitive constructions of Voronoi diagrams~\citep{nielsen2010hyperbolic}, Delaunay graphs~\citep{medbouhi2024hyperbolic} and SVM decision boundaries~\citep{cho2019large}. \citet{celinska2024numerical} studied numerical precision of hyperbolic models including the Klein model. Moreover, properties of the Klein model are also used to assist in proofs for other hyperbolic models~\citep{he2024lorentzian}. However, the above works did not explore neural operations in the Klein model.

Although early works~\citep{taherian2010algebraic, rostamzadeh2014trigonometry} studied the algebraic structures and trigonometry of the Klein model, they did not address constructing neural network layers. While \citet{ungar2022gyrovector, ungar2012beyond, kim2013unit} summarized the gyrogroup properties of the Einstein operations in connection with the Klein model, they did not mention the properties that we introduce in this paper.

\section{The Klein Model of Hyperbolic Geometry}\label{sec:klein_model}

The $n$-dimensional hyperbolic space, denoted as $\sH^n$, is a simply connected $n$-dimensional Riemannian manifold with constant negative curvature. To represent points in $\sH^n$, there are several isometric models including the Poincar\'e ball model $\sB^n$ and the hyperboloid (Lorentz) model $\sL^n$, both of which are commonly used in the HNN literature. The Klein model (also known as Beltrami-Klein), denoted as $\sK^n$, is another notable representation. In our paper, we consider $\sH^n$ with a curvature $-1$ since this is used in most of the HNN literature.

Recent works have derived compact formulas for basic operations in $\sB^n$ and $\sL^n$~\citep{ganea2018hyperbolic, chami2019hyperbolic}. Our derivation will use these formulas to avoid complicated calculations of, e.g., connections. In this section, we introduce the Klein model and describe the isometric mappings between the Klein model and the other two models. A detailed review of $\sB^n$ and $\sL^n$ is provided in \appendixref{apd:first}. 

The $n$-dimensional Klein model with a curvature $-1$ is represented as $\sK^n = \{\rvx \in \sR^n, \|\rvx\| < 1\}$, with the Riemannian metric tensor $g_\rvx^\gK=g^\gK(\rvx)$ expressed as:
\begin{equation}
    g_{ij}^\gK(\rvx) = \frac{\delta_{ij}}{1-\norm{\rvx}^2} + \frac{x_i x_j}{( 1 - \norm{\rvx}^2 )^2},
\end{equation}
where $\norm{\cdot}$ is the Euclidean norm. 
Denoting points in $\sR^{n+1}$ as $[x_0,x_1,\cdots,x_n]^\T$, the Klein ball can be obtained by mapping $\rvx \in \sL^n$ to the hyperplane $x_0 = 1$, using a gnomonic (central) projection with rays emanating from the origin. Specifically, we review the following formulas for isometric mappings between hyperbolic models. We say two points are ``corresponding'' if they are related by such an isometric mapping.

\paragraph{Mappings between models}
We use $\pi_{\gL \to \gK}$ to denote the isometric mapping from $\sL^n$ to $\sK^n$ and $\pi_{\gK \to \gL}$ the isometric mapping from $\sK^n$ to $\sL^n$. Namely, for $\rvx^\gL = [x_0^\gL, x_1^\gL, \cdots, x_n^\gL]^\T \in \sL^n$ and $\rvx^\gK = [x_1^\gK, \cdots, x_n^\gK]^\T \in \sK^n$, 
\begin{equation}\label{eq:iso_map_L_K}
\pi_{\gL\to\gK}(\rvx^\gL) = \left[\frac{x_1^\gL}{x_0^\gL}, \cdots, \frac{x_n^\gL}{x_0^\gL}\right]^\T \in \sK^n, \quad \pi_{\gK\to\gL}(\rvx^\gK) = \frac{1}{\sqrt{1-\|\rvx^\gK\|^2}}\left[1, \rvx_\gK^\T \right]^\T \in \sL^n.
\end{equation}

We use $\pi_{\gB \to \gK}$ to denote the isometric mapping from $\sB^n$ to $\sK^n$ and $\pi_{\gK \to \gB}$ the isometric mapping from $\sK^n$ to $\sB^n$. Namely, for $\rvx^\gB \in \sB^n$ and $\rvx^\gK \in \sK^n$, 
\begin{equation}\label{eq:iso_map_B_K}
\pi_{\gB\to\gK}(\rvx^\gB) = \frac{2}{1+\|\rvx^\gB\|^2}\rvx^\gB \in \sK^n, \quad \pi_{\gK\to\gB}(\rvx^\gK) = \frac{1}{1+\sqrt{1-\|\rvx^\gK\|^2}}\rvx^\gK \in \sB^n.
\end{equation}

In $\sK^n$, the tangent space $\gT_{\rvx^\gK}\sK^n$ at $\rvx^\gK$ is represented as $\sR^n$ where $\frac{\partial}{\partial x_i^\gK}$ is represented as $\rve_i$, the $i$-th canonical basis. Similar conventions are adopted for $\sB^n$ and $\sL^n$. The tangent vectors are related using pushforward maps associated with the above isometric mappings. We say that two vectors are ``corresponding'' if they are related by such a pushforward map. We recall that a pushforward map of an isometry preserves inner products and parallel transports.

The following lemma presents formulas for obtaining corresponding tangent vectors between $\sK^n$ and other models. The proofs of all the results are presented in \appendixref{apd:second}.
 
\begin{lemma}\label{lem:tangent vectors in Klein and Poincare}
Let $\rvx^\gK \in \sK^n$ and $\rvv^\gK \in \gT_{\rvx^\gK} \sK^n$ be a tangent vector at $\rvx^\gK$.
\begin{enumerate}
    \item Let $\rvx^\gB \in \sB^n$ be the corresponding point of $\rvx^\gK$ and $\rvv^\gB \in \gT_{\rvx^\gB}\sB^n$ be the corresponding tangent vector of $\rvx^\gK$. Then
    \begin{align}
    \rvv^\gK &= \frac{\partial \pi_{\gB\to\gK}(\rvx^\gB)}{\partial \rvx^\gB} \rvv^\gB = \frac{2}{1+\|\rvx^\gB\|^2} \rvv^\gB - \frac{4\rvx^\gB\cdot\rvv^\gB}{(1+\|\rvx^\gB\|^2)^2}\rvx^\gB , \label{eq:yidan_review_b1} \\
    \rvv^\gB &= \frac{\partial \pi_{\gK\to\gB}(\rvx^\gK)}{\partial \rvx^\gK} \rvv^\gK \nonumber \\
    &= \frac{1}{1+\sqrt{1-\|\rvx^\gK\|^2}} \rvv^\gK +\frac{\rvx^\gK\cdot\rvv^\gK}{\sqrt{1-\|\rvx^\gK\|^2}(1+\sqrt{1-\|\rvx^\gK\|^2})^2}  \rvx^\gK . \label{eq:yidan_review_b2} 
    \end{align}
    \item Let $\rvx^\gL \in \sL^n$ be the corresponding point of $\rvx^\gK$ and $\rvv^\gL \in \gT_{\rvx^\gL}\sL^n$ be the corresponding tangent vector of $\rvx^\gK$. Then
    \begin{align}
    \rvv^\gK &= \frac{\partial \pi_{\gL\to\gK}(\rvx^\gL)}{\partial \rvx^\gL} \rvv^\gL = -\frac{v_t}{x_t^2}\rvx_s + \frac{1}{x_t}\rvv_s , \label{eq:yidan_review_1}\\
    \rvv^\gL &= \frac{\partial \pi_{\gK\to\gL}(\rvx^\gK)}{\partial \rvx^\gK} \rvv^\gK = \begin{bmatrix}\frac{\rvx^\gK\cdot\rvv^\gK}{\left(1-\|\rvx^\gK\|^2\right)^{\frac{3}{2}}}\\\frac{1}{\sqrt{1-\|\rvx^\gK\|^2}}\rvv^\gK + \frac{\rvx^\gK\cdot \rvv^\gK }{\left(1-\|\rvx^\gK\|^2\right)^{\frac{3}{2}}} \rvx^\gK \end{bmatrix} . \label{eq:yidan_review_2}
    \end{align}
\end{enumerate}
\end{lemma}


\paragraph{Distances} The induced distance function on the Klein model can be inferred from the one on the hyperboloid model, which we present as the following lemma.
\begin{lemma}\label{pro:distance}
Let $\rvx^\gK, \rvy^\gK \in \sK^n$. Their geodesic distance is given by
\begin{equation}
d_{\gK} (\rvx^\gK, \rvy^\gK) = d_{\gL} (\pi_{\gK \rightarrow \gL}(\rvx^\gK), \pi_{\gK \rightarrow \gL}(\rvy^\gK)) = \cosh^{-1} \left( \frac{1 - \rvx^\gK\cdot \rvy^\gK }{\sqrt{1-\| \rvx^\gK \|^2} \sqrt{1-\| \rvy^\gK \|^2}}  \right).
\end{equation}
\end{lemma}

\paragraph{Unit-speed geodesics} The parametric expression of unit-speed geodesics in the Klein model is given in the following lemma, which is derived based on the fact that an isometric mapping maps geodesics in $\sL^n$ to geodesics in $\sK^n$ .
\begin{lemma}\label{lem:geodesic}
Let $\rvx^\gK \in \sK^n$ and $\rvv^\gK \in \gT_\rvx \sK^n$ with $g_{\rvx^\gK}^{\gK}( \rvv^\gK, \rvv^\gK) = 1$. Let $\gamma_{\rvx^\gK,\rvv^\gK}(t)$ denote the unit-speed geodesic in $\sK^n$ with $\gamma_{\rvx^\gK,\rvv^\gK}(0) = \rvx^\gK$ and $\dot{\gamma}_{\rvx^\gK,\rvv^\gK}(0) = \rvv^\gK$, then
\begin{equation}
\gamma_{\rvx^\gK,\rvv^\gK}(t) = \rvx^\gK + \frac{\sinh(t) \rvv^\gK} {\cosh(t) + \lambda_{\rvx^\gK} (\rvx^\gK\cdot\rvv^\gK) \sinh(t)},
\end{equation}
where $\lambda_{\rvx^\gK} := 1/\sqrt{1-\| \rvx^\gK \|^2}$ is used globally in this paper.
\end{lemma}

Accordingly, we can write out the exponential maps and logarithmic maps for the Klein model, summarized as the following corollary.

\begin{corollary}\label{cor:exp and log maps}
Given $\rvx^\gK \in \sK^n$ and $\rvv^\gK \in \gT_{\rvx^\gK} \sK^n$, denote $\rvu^\gK = \rvv^\gK / \| \rvv^\gK \|_{\gK}$ with $\| \cdot \|_{\gK}^2 = { g_{\rvx^\gK}^{\gK} (\cdot, \cdot) }$. Let $\gamma_{\rvx^\gK,\rvu^\gK}(t)$ be the unit-speed geodesic in $\sK^n$ with $\gamma_{\rvx^\gK,\rvu^\gK}(0) = \rvu^\gK$ and $\dot{\gamma}_{\rvx^\gK,\rvu^\gK}(0) = \rvu^\gK$, the exponential map $\exp^{\gK}_{\rvx^\gK} : \gT_{\rvx^\gK} \sK^n \rightarrow \sK^n$ is given by
\begin{equation}\label{eq:exp_K}
\exp^{\gK}_{\rvx^\gK} (\rvv^\gK) = \gamma_{\rvx^\gK,\rvu^\gK}(\| \rvv^\gK \|_{\gK}) = \rvx^\gK + \frac{\sinh(\| \rvv^\gK \|_{\gK}) \frac{\rvv^\gK}{\| \rvv^\gK \|_{\gK}}} {\cosh(\| \rvv^\gK \|_{\gK}) + \frac{\lambda^2_{\rvx^\gK}}{\| \rvv^\gK \|_{\gK}} (\rvx^\gK\cdot\rvv^\gK)\sinh(\| \rvv^\gK \|_{\gK})}.
\end{equation}
The logarithmic map $\log^{\gK}_{\rvx^\gK} : \sK^n \rightarrow \gT_{\rvx^\gK} \sK^n$ is given by 
\begin{equation}\label{eq:log_K}
\log^{\gK}_{\rvx^\gK}(\rvy^\gK) = d_{\gK} (\rvx^\gK, \rvy^\gK) \frac{\rvy^\gK-\rvx^\gK}{\| \rvy^\gK-\rvx^\gK \|}_{\gK} = \cosh^{-1} \left( \frac{1 -\rvx^\gK\cdot\rvy^\gK}{\sqrt{1-\| \rvx^\gK \|^2} \sqrt{1-\| \rvy^\gK \|^2}} \right) \frac{\rvy^\gK-\rvx^\gK}{\| \rvy^\gK-\rvx^\gK \|}_{\gK},
\end{equation}
In particular, it follows that
\begin{equation}\label{eq:exp_log_K_0}
\exp^{\gK}_{\rvo^\gK} (\rvv^\gK) = \tanh(\| \rvv^\gK \|) \frac{\rvv^\gK}{\| \rvv^\gK \|}, \quad \log^{\gK}_{\rvo^\gK}(\rvy^\gK) = \cosh^{-1} (\lambda_{\rvy^\gK}) \frac{\rvy^\gK}{\| \rvy^\gK\|}.
\end{equation}
\end{corollary}

\paragraph{Parallel transport} The parallel transport $P_{\rvx\to\rvy}$ defines a linear isometry moving tangent vectors along the geodesic from $\rvx$ to $\rvy$. Unfortunately, the formula for parallel transport between general points would be complicated in the Klein model. We only present the case where the starting point is $\rvo^\gK$, the hyperbolic origin in $\sK^n$, which is sufficient for the use case of hyperbolic neural networks.



\begin{proposition}\label{pro:parallel transport in Klein}
The parallel transport of a tangent vector $\rvv^\gK\in\gT_{\rvo^\gK}\sK^n$ to the tangent space $\gT_{\rvx^\gK}\sK^n$ at an arbitrary point $\rvx^\gK$ in the Klein model is
\begin{equation}
P_{\rvo^{\gK}\to\rvx^{\gK}}(\rvv^{\gK}) = \frac{(\rvx^\gK\cdot\rvv^\gK)(\sqrt{1-\|\rvx^\gK\|^2}-2)}{1-\sqrt{1-\|\rvx^\gK\|^2}}\rvx^\gK +\sqrt{1-\|\rvx^\gK\|^2} \rvv^\gK\in \gT_{\rvx^\gK}\sK^n.
\end{equation}
\end{proposition}

\section{Klein Model for Hyperbolic Neural Networks}
To build HNNs in hyperbolic space, we need to define operations including weight matrix transformation and bias translation in a way analogous to ``$\rmW \rvx + \rvb$''. A natural approach is as follows. First, apply the logarithmic map to project the input to the tangent space at the hyperbolic origin. Then for weight matrix transformation, perform the matrix-vector multiplication in the tangent space; for bias translation,  apply the parallel transport to the hyperbolic bias vector from the origin to the input point. Finally, use the exponential map to bring the output back to hyperbolic space. For the Poincar\'e ball model, \citet{ganea2018hyperbolic} proved that these ``tangent space operations'' are equivalent to basic operations in the M\"obius gyrovector spaces. In the context of the Klein model, the corresponding gyrovector space is the Einstein gyrovector space. This raises a natural question: are the tangent space operations in the Klein model also equivalent to those in the Einstein gyrovector space?

In \sectionref{sec:ein_gyro_space}, we review the definitions of the Einstein gyrovector spaces. We prove the equivalence of tangent space operations and Einstein gyrovector space operations in \sectionref{sec:conn_geo_gyro}. Finally, we derive HNNs in the Klein model in \sectionref{sec:hnn_klein}.

\subsection{Einstein Gryogroups and Gyrovector Spaces}\label{sec:ein_gyro_space}

\paragraph{Einstein gyrogroup}
Let $\sV$ be a real inner product space and let $\sV_c = \{\rvx\in\sV: \|\rvx\|<c\}$ be an open $c$-ball of $\sV$. Then the Einstein addition of $\rvx, \rvy \in \sV_c$ is defined as 
\begin{equation}
\rvx\oplus_\text{E}\rvy=\frac1{1+\frac{\rvx\cdot\rvy}{c^2}}\left(\rvx+\frac1{\gamma_\rvx}\rvy+\frac{1}{c^2}\frac{\gamma_\rvx}{1+\gamma_\rvx}(\rvx\cdot\rvy)\rvx\right),
\label{eq:ein-add}
\end{equation}
where $\gamma_\rvx=1/\sqrt{1-\tfrac{\|\rvx\|^2}{c^2}}$ is the Lorentz factor. These designations originate from Einstein's theory of special relativity: if $\sV=\R^3$, \equationref{eq:ein-add} represents the relativistic addition of velocities $\rvx\,,\rvy$ in space rather than spacetime, where $c$ is the speed of light.

In the general case where $\rvx\,,\rvy$ are not parallel, $\oplus_\text{E}$ is neither commutative nor associative, rendering the corresponding algebraic structure merely a groupoid. As observed by Ungar, however, this groupoid has the interesting property whereby an automorphism $\mathrm{gyr}[\rvx\,,\rvy]:\sV_c\rightarrow\sV_c$ respecting $\oplus_E$, called gyration, defines the obstruction to commutativity through $\rvx\oplus_\text{E}\rvy=\mathrm{gyr}[\rvx\,,\rvy]\rvy\oplus_\text{E}\rvx$ as well as to associativity through $\rvx\oplus_\text{E}(\rvy\oplus_\text{E}\rvz)=(\rvx\oplus_\text{E}\rvy)\oplus_\text{E}\mathrm{gyr}[\rvx\,,\rvy]\rvz$.
This groupoid structure $(\sV_c\,,\oplus_\text{E})$ is called a gyrocommutative gyrogroup, and describes Thomas precession in the example of special relativity. We refer to $(\sV_c\,,\oplus_\text{E})$ as Einstein gyrogroup for short, and more details on its properties can be found, e.g., in \citet[sec. 1.2, 2.3]{ungar2022gyrovector}.

\paragraph{Einstein gyrovector space}
Given an Einstein gyrogroup, we can endow it with another operation called Einstein scalar multiplication: letting $r\in\sR, \rvx\in\sV_c$, and $\rvx\neq \mathbf{0}$, it is
\begin{equation}
r\otimes_\text{E} \rvx = c\tanh{\left(r\tanh^{-1}\left(\frac{\|\rvx\|}{c}\right)\right)}\frac{\rvx}{\|\rvx\|}\,.
\end{equation}
This turns out to have associative and distributive properties, and we apply the notation $r\otimes_\text{E}\rvx=\rvx\otimes_\text{E}r$. 
The resulting structure $(\sV_c\,, \oplus_\text{E}\,,\otimes_\text{E})$ is called Einstein gyrovector space, and see, e.g., \citet[sec. 3.1, 3.8]{ungar2022gyrovector} for relevant properties. 
Notably, its inherited inner product is invariant under gyrations, that is, $\mathrm{gyr}[\rvx\,,\rvy]\rvu\cdot \mathrm{gyr}[\rvx\,,\rvy]\rvv=\rvu\cdot\rvv$ for all $\rvu\,,\rvv \in \sV_c$. 
Einstein gyrovector space turns out to be a useful framework for hyperbolic geometry, since we recover the Klein model of curvature $-1$ for $\sV = \R^n$ and $c=1$.

\subsection{Connecting the Klein Model with Einstein Gyrovector Spaces}\label{sec:conn_geo_gyro}

Next, we show how Einstein gyrovector spaces facilitate the tangent space construction of scalar multiplication and vector addition in the Klein model. 
We start with the geodesics in the Klein model and their formulation using operations in the Einstein gyrovector spaces.

\paragraph{Geodesics} The geodesic $\gamma_{\rvx^\gK\to\rvy^\gK}: \sR\to\sK^n$ connecting points $\rvx^\gK, \rvy^\gK \in\sK^n$, such that  $\gamma_{\rvx^\gK\to\rvy^\gK}(0)=\rvx^\gK$ and $\gamma_{\rvx^\gK\to\rvy^\gK}(1)=\rvy^\gK$, is shown by \citet[sec. 3.9]{ungar2022gyrovector} to be:
\begin{equation}
\gamma_{\rvx^\gK\to\rvy^\gK}(t) = \rvx^\gK\oplus_\text{E}\left(\ominus_\text{E}\rvx^\gK\oplus_\text{E}\rvy^\gK\right)\otimes_\text{E}t,
\end{equation}
where $\ominus_\text{E} \rvx^\gK$ is the inverse of $\rvx^\gK$ in the Einstein gyrogroup: $\ominus_\text{E} \rvx^\gK  \oplus_\text{E} \rvx^\gK = \rvo^\gK$.

\paragraph{Scaling and Einstein scalar multiplication} We study the tangent space construction of performing scalar multiplication in the Klein model and build the identity with the Einstein addition. The following theorem summarizes our finding: 

\begin{theorem}\label{thm:Klein and Einstein scalar multiplication}
For $\forall r \in \sR, \rvx^\gK\in\sK^n$, performing scalar multiplication in the Klein model, which first uses the logarithmic map to project $\rvx^\gK$ to $\gT_{\rvo^\gK}\sK^n$, then multiplies this projection by a scalar in the tangent space and projects it back to the manifold with the exponential map, can be achieved by directly applying the Einstein scalar multiplication, namely, 
\begin{equation}\label{eq:Klein and Einstein scalar multiplication}
r\otimes_\text{E} \rvx^\gK = \exp^\gK_{\rvo^\gK}\left(r\log^\gK_{\rvo^\gK}\left(\rvx^\gK\right)\right).
\end{equation}
\end{theorem}


\paragraph{Parallel transport and Einstein addition} We also connect the parallel transport with the Einstein addition using exponential and logarithmic maps in the following theorem:

\begin{theorem}\label{thm:Klein and Einstein addition}
The parallel transport w.r.t. the Levi-Civita connection of a tangent vector $\rvv^\gK \in \gT_{\rvo^\gK}\sK^n$ to another tangent space $\gT_{\rvx^\gK}\sK^n$ in the Klein model is given by
\begin{equation}\label{eq:yidan_pt_K}
P_{\rvo^{\gK}\to\rvx^{\gK}}(\rvv^{\gK}) = \log^\gK_{\rvx^\gK}\left(\rvx^{\gK}\oplus_\text{E}\exp^\gK_{\rvo^\gK}(\rvv^{\gK})\right).
\end{equation}
\end{theorem}


\subsection{Hyperbolic Neural Networks in the Klein Model}\label{sec:hnn_klein}

Theorems~\ref{thm:Klein and Einstein scalar multiplication} and \ref{thm:Klein and Einstein addition} 
provide compact formulas for defining neural operations for HNNs using the Klein model.

\paragraph{Weight matrix transformation and nonlinear activation}
First, thanks to \theoremref{thm:Klein and Einstein scalar multiplication}, we define the ``Einstein version'' of Euclidean functions that preserve the origin.


\begin{definition}[Einstein version]\label{def:einstein version}
Given $f: \sR^n \to \sR^m$ such that $f(\vzero) = (\vzero)$, the Einstein version of $f$, denoted as $f^{\otimes_\text{E}}: \sK^n \to \sK^m$, is defined by 
\begin{equation}
f^{\otimes_\text{E}}(\rvx^\gK):=\exp^{\gK}_{\rvo^\gK}\left(f\left(\log^{\gK}_{\rvo^\gK}(\rvx^\gK)\right)\right).
\end{equation}
\end{definition}

For instance, if $\sigma: \sR^n\to\sR^n$ is a non-linear activation function, then its Einstein version $\sigma^{\otimes_\text{E}}$ can be applied to points in the Klein model. Namely, 
\begin{equation}
\sigma^{\otimes_\text{E}}(\rvx^\gK):=\exp^{\gK}_{\rvo^\gK}\left(\sigma\left(\log^{\gK}_{\rvo^\gK}(\rvx^\gK)\right)\right).
\end{equation}
Moreover, we identify matrix-vector multiplication as an Einstein version of linear maps. Then the following result is straightforward.

\begin{theorem}[Einstein matrix-vector multiplication]\label{thm:Einstein matrix-vector multiplication} 
 If $\rmM:\sR^n \to \sR^m$ is a linear map, which we identify with its matrix representation, then for $\rvx^\gK\in\sK^n$ s.t. $\rmM\rvx^\gK \neq \mathbf{0}$,
\begin{equation}
\begin{aligned}
\rmM^{\otimes_\text{E}}(\rvx^\gK) &= \exp^\gK_{\rvo^\gK}(\rmM\log^\gK_{\rvo^\gK}( \rvx^\gK)) \\
&= \tanh \left(\frac{2\|\rmM \rvx^\gK\|}{\|\rvx^\gK\|} \tanh ^{-1}\left(\frac{\|\rvx^\gK\|}{1+\sqrt{1-\|\rvx^\gK\|^2}}\right)\right) \frac{\rmM \rvx^\gK}{\|\rmM\rvx^\gK\|};
\end{aligned}
\end{equation}
and if $\rmM\rvx^\gK = \mathbf{0}$, then $\rmM^{\otimes_\text{E}}(\rvx^\gK) = \mathbf{0}$.
Moreover, if we define the Einstein matrix-vector multiplication of $\rmM\in\gM_{m,n}(\sR)$ and $\rvx^\gK\in\sK^n$ by $\rmM\otimes_\text{E}\rvx^\gK:=\rmM^{\otimes_\text{E}}(\rvx^\gK)$, then
it shares common properties with the matrix-vector multiplication in the Euclidean space.\\
1. For $\rmM \in \gM_{l,m}(\sR)$, $\rmM' \in \gM_{m,n}(\sR)$,    
$(\rmM\rmM') \otimes_\text{E} \rvx^\gK = \rmM \otimes_\text{E} (\rmM' \otimes_\text{E} \rvx^\gK)$; \\ 
2. For $r>0$, $\rmM \in \gM_{m,n}(\sR)$,
$(r\rmM) \otimes_\text{E} \rvx^\gK = r \otimes_\text{E} (\rmM \otimes_\text{E} \rvx^\gK)$; \\ 
3. For $\rmM \in \gO_n(\sR)$, $\rmM \otimes_\text{E} \rvx^\gK = \rmM \rvx^\gK$. 
\end{theorem}


\paragraph{Bias translation} 
Next, \theoremref{thm:Klein and Einstein addition} provides a compact formula for bias translation. Specifically, we can use the Einstein addition in place of the bias translation of a point $\rvx^{\gK}\in\sK^n$ with a hyperbolic bias $\rvb^{\gK}\in\sK^n$, 
\begin{equation}
\begin{aligned}
\rvx^{\gK} &\mapsto \exp^\gK_{\rvx^\gK}\left(P_{\rvo^{\gK}\to\rvx^{\gK}}^{\gK}\left(\log^\gK_{\rvo^\gK}(\rvb^{\gK})\right)\right)\\
&=\exp^\gK_{\rvx^\gK}\left(\log^\gK_{\rvx^\gK}\left(\rvx^{\gK}\oplus_\text{E}\exp^\gK_{\rvo^\gK}\left(\log^\gK_{\rvo^\gK}\left(\rvb^{\gK}\right)\right)\right)\right)=\rvx^{\gK}\oplus_\text{E}\rvb^{\gK}.
\end{aligned}
\end{equation}

In summary, to build Klein HNNs, Einstein operations are all we need.

\section{Experiments}
We perform node classification tasks on the well-known WebKB datasets including Texas, Wisconsin and Chameleon~\citep{craven2000learning}, Actor datasets~\citep{tang2009social} and citation datasets including Cora and Pubmed~\citep{sen2008collective}, summarized in Appendix~\ref{apd:third}. These datasets enjoy high Gromov hyperbolicities~\citep{jonckheere2008scaled, adcock2013tree}.
We compare HNNs using the three hyperbolic models we have discussed. We use HNNs containing a hyperbolic linear layer and a Euclidean classification layer. We use (hyperbolic version of) ReLU as the activation function. The HNNs are trained using the Riemannian Adam~\citep{becigneul2018riemannian} for 5000 epochs with early stopping.

\tableref{tab:test accuracy} presents the test accuracies. The results indicate that Klein HNNs performs on par with both HNNs using the Poincar\'e ball model and the hyperboloid model.

\begin{table}[hbtp]
\floatconts
{tab:test accuracy}
  \centering
  {\caption{The test accuracy of two-layer HNNs (mean and std over 3 trials)}\label{tab:test accuracy}}
  \tiny
  {  \begin{tabular}{l|llllll}
  \toprule
  \bfseries Dataset & \bfseries Texas & \bfseries Wisconsin & \bfseries Chameleon & \bfseries Actor & \bfseries Cora  &\bfseries Pubmed\\
  \midrule
  Poincar\'e  & 0.9697±0.0347   & 0.9506±0.0283   & 0.7442±0.0064  & 0.6436±0.0057   &  0.5960±0.0101 & 0.7270±0.0036  \\
Hyperboloid & 0.9697±0.0132   & 0.9444±0.0321   & 0.7418±0.0120  & 0.6342±0.0226   & 0.6067±0.0137 &0.7293±0.0064 \\
Klein     & 0.9697±0.0263  & 0.9568±0.0214 & 0.7375±0.0119 & 0.6509±0.0067    &  0.5957±0.0106&0.7230±0.0139\\
  \bottomrule
  \end{tabular}}
\end{table}

In \tableref{tab:runtime}, we report the average run time to train each epoch. Evidently, the average training time per epoch for the hyperboloid HNN is longer than those for Poincar\'e ball and Klein HNNs. This is because no simple formulation exists for the weight matrix transformation and bias translation in the hyperboloid model and we have to perform the tangent space operations which involve a series of exponential and logarithmic maps, which cause computational complexity. Thanks to the connection between the Klein model and Einstein gyrovector spaces, we are able to implement the Klein linear layer in a simple and elegant way. Emperically, Klein HNNs are as efficient as Poincar\'e HNNs and sometimes the most efficient among all the three models.

\begin{table}[hbtp]
\floatconts
  {tab:runtime}
  \centering
  {\caption{Average runtime for training one epoch (mean and std over 3 trials)}\label{tab:runtime}}
  \tiny
  {\begin{tabular}{l|llllll}
  \toprule
  \bfseries Dataset & \bfseries Texas & \bfseries Wisconsin & \bfseries Chameleon & \bfseries Actor & \bfseries Cora  &\bfseries Pubmed\\
  \midrule
Poincar\'e  & 0.0222±0.0025  & 0.0205±0.0008   & 0.0248±0.0008  & 0.0263±0.0009   & 0.0198±0.0010 & 0.0220±0.0023\\
Hyperboloid & 0.0396±0.0087   & 0.0341±0.0012   &  0.0224±0.0021   & 0.0420±0.0014   & 0.0362±0.0022 & 0.0351±0.0014\\
Klein     & 0.0219±0.0017  & 0.0208±0.0020 & 0.0234±0.0029 & 0.0238±0.0008  &  0.0214±0.0012 &0.0222±0.0017\\
  \bottomrule
  \end{tabular}}
\end{table}

To study how training loss changes differently with training epochs using different models, we plot their relationships when training on some selected datasets. \figureref{fig:train_loss_change} shows the result on the Wisconsin, Cora, and Pubmed datasets. We observe that the training dynamics of the HNNs are very similar.

\begin{figure}[htbp]
\floatconts
  {fig:train_loss_change}
  {\caption{Change of training loss with epochs using different models.}}
  {%
    \subfigure[Wisconsin]{\label{wisconsin_loss}%
      \includegraphics[width=0.29\linewidth]{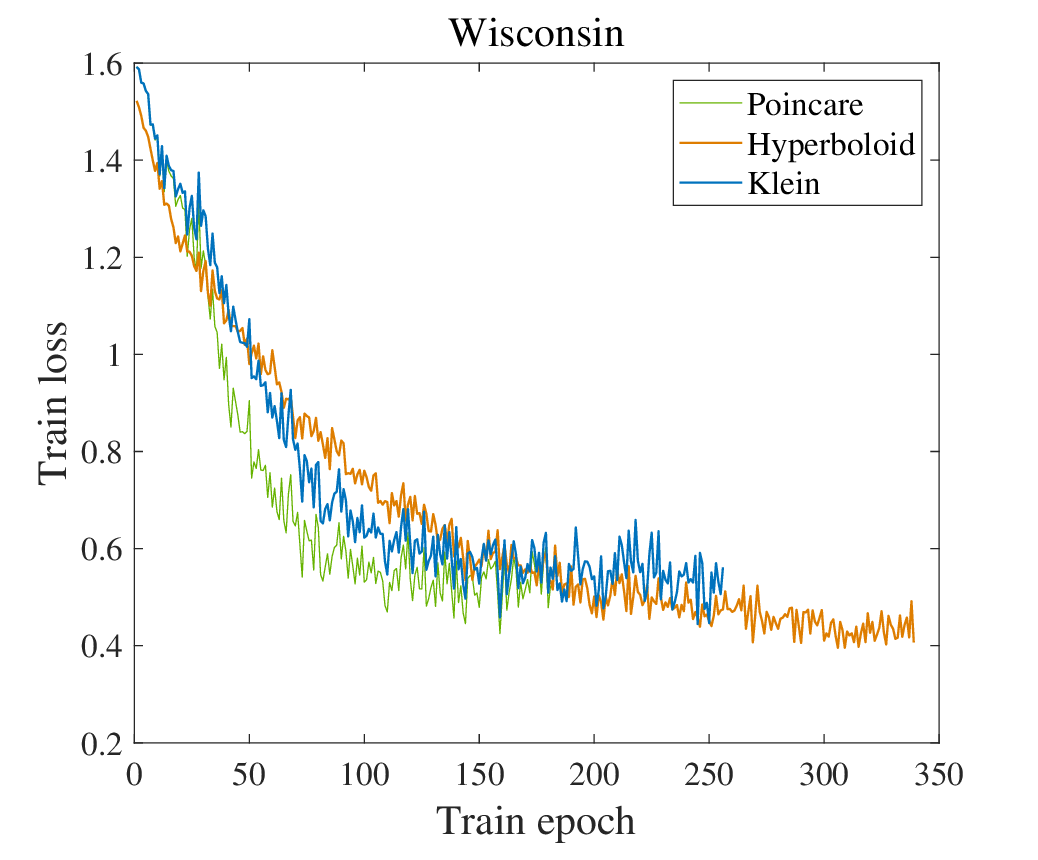}}%
    \qquad
    \subfigure[Cora]{\label{cora_loss}%
      \includegraphics[width=0.29\linewidth]{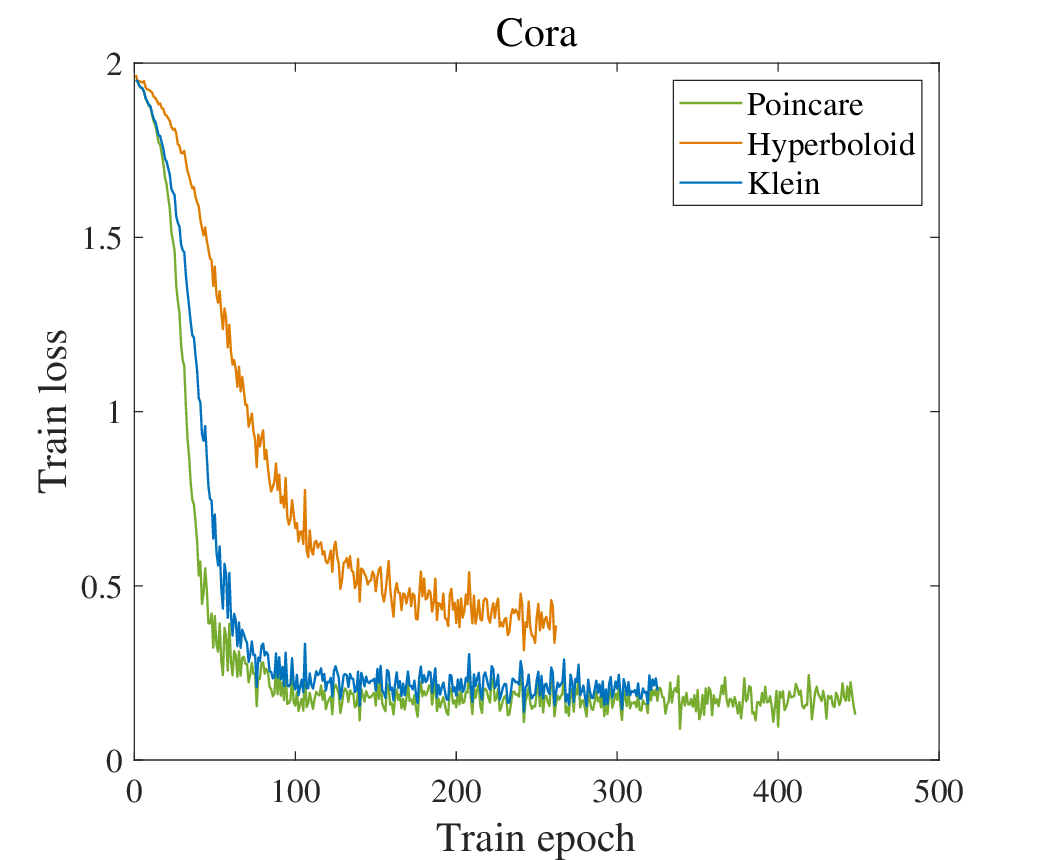}}%
    \qquad
    \subfigure[Pubmed]{\label{pubmed_loss}%
      \includegraphics[width=0.29\linewidth]{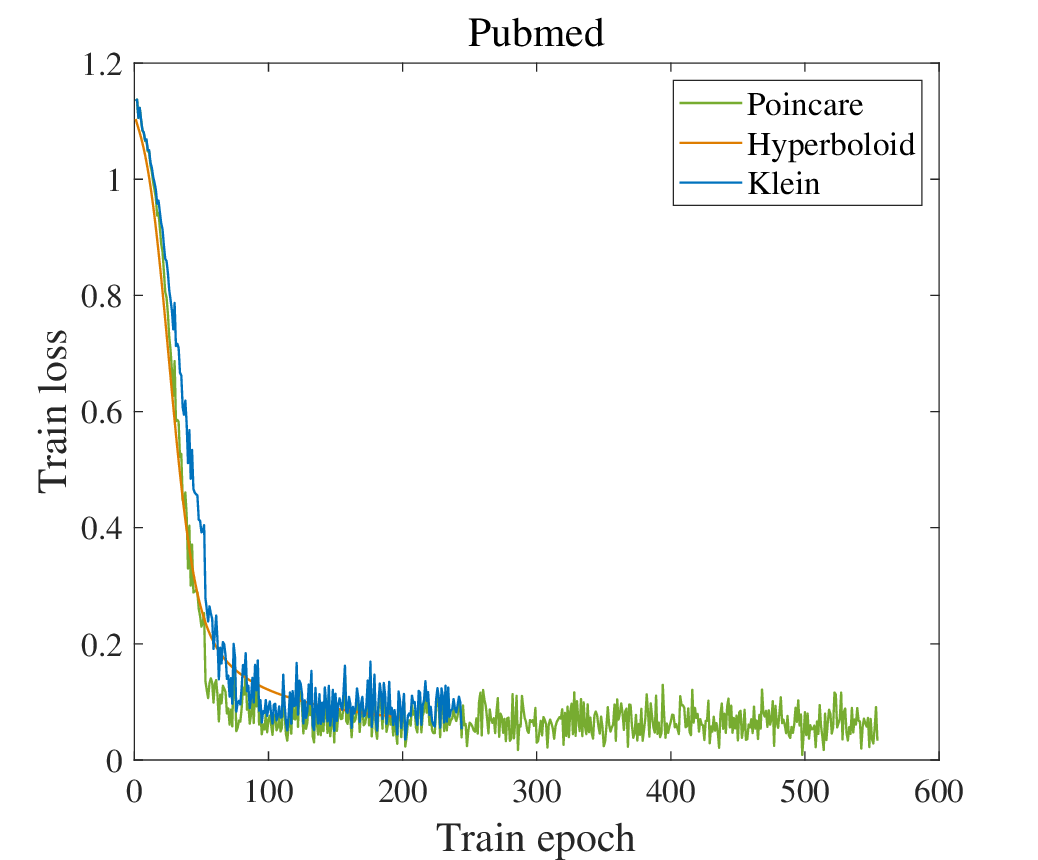}}
  }
\end{figure}

We also plot the hyperbolic features, i.e., output of the hyperbolic linear layer. For illustration, we project them into the 2D-disk using hyperbolic t-SNE. \figureref{fig:features2d} shows the results on the Texas and Wisconsin datasets for Poincar\'e and Klein HNNs. Features with different labels are indicated by different colors. We observe that the representations show classification results for the same dataset, indicating the effectiveness of our Klein HNNs. 

\begin{figure}[htbp]
\floatconts
  {fig:features2d}
  {\caption{Features projected into the 2D-disks.}}
  {%
    \subfigure[Texas ($\sB$)]{\label{texas_poincare}%
      \includegraphics[width=0.21\linewidth]{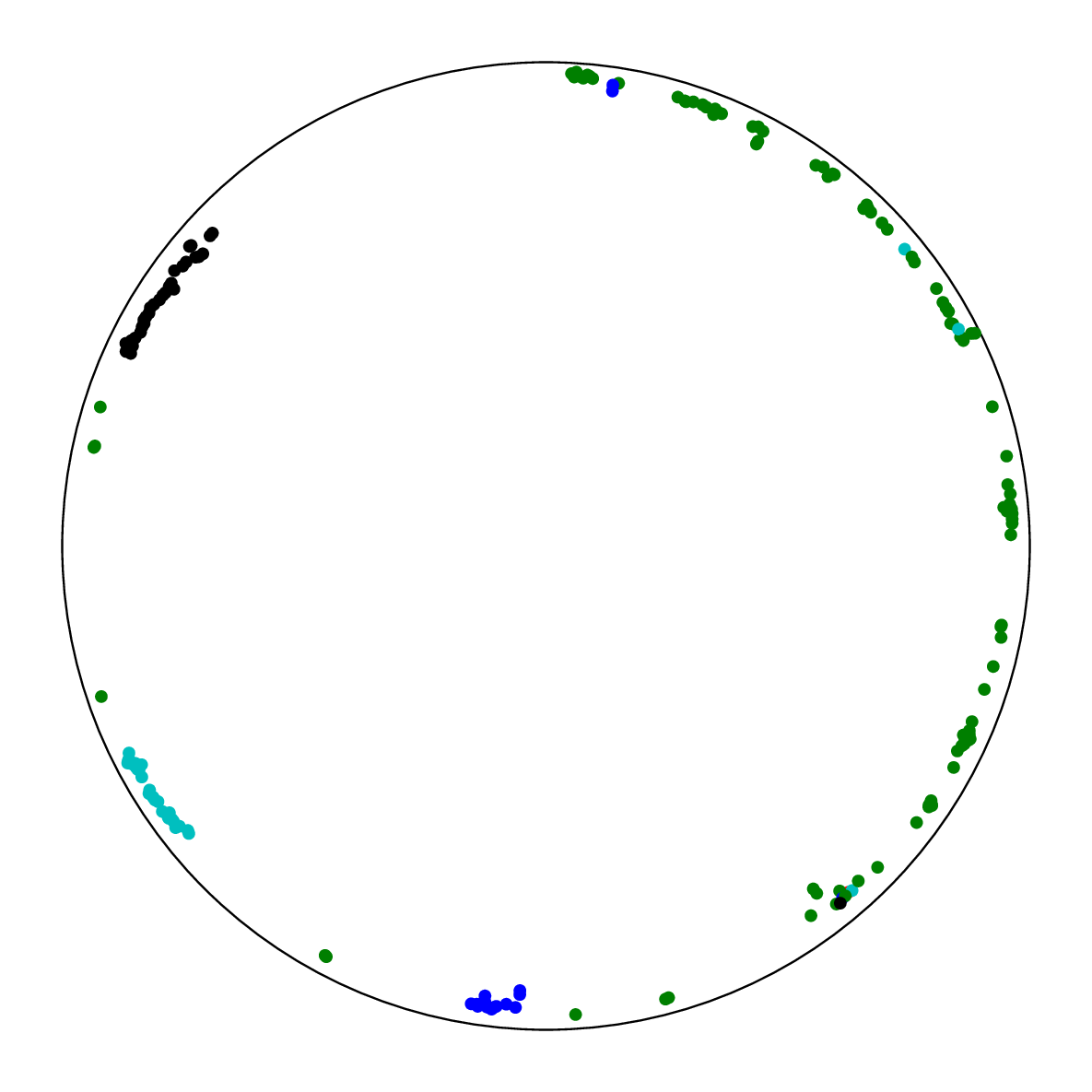}}%
    \qquad
    \subfigure[Texas ($\sK$)]{\label{texas_klein}%
      \includegraphics[width=0.21\linewidth]{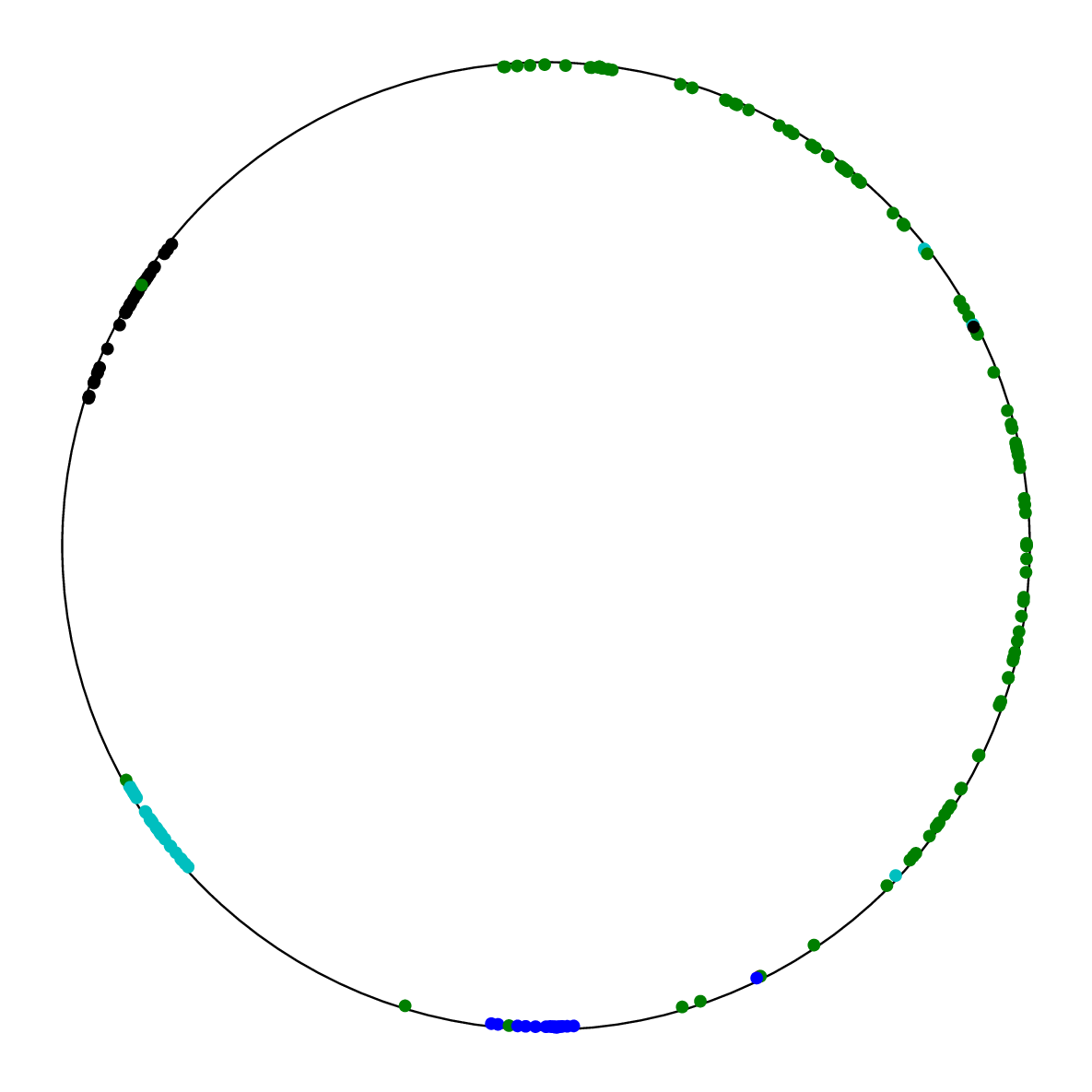}}%
    \qquad
    \subfigure[Wisconsin ($\sB$)]{\label{wisconsin_poincare}%
      \includegraphics[width=0.21\linewidth]{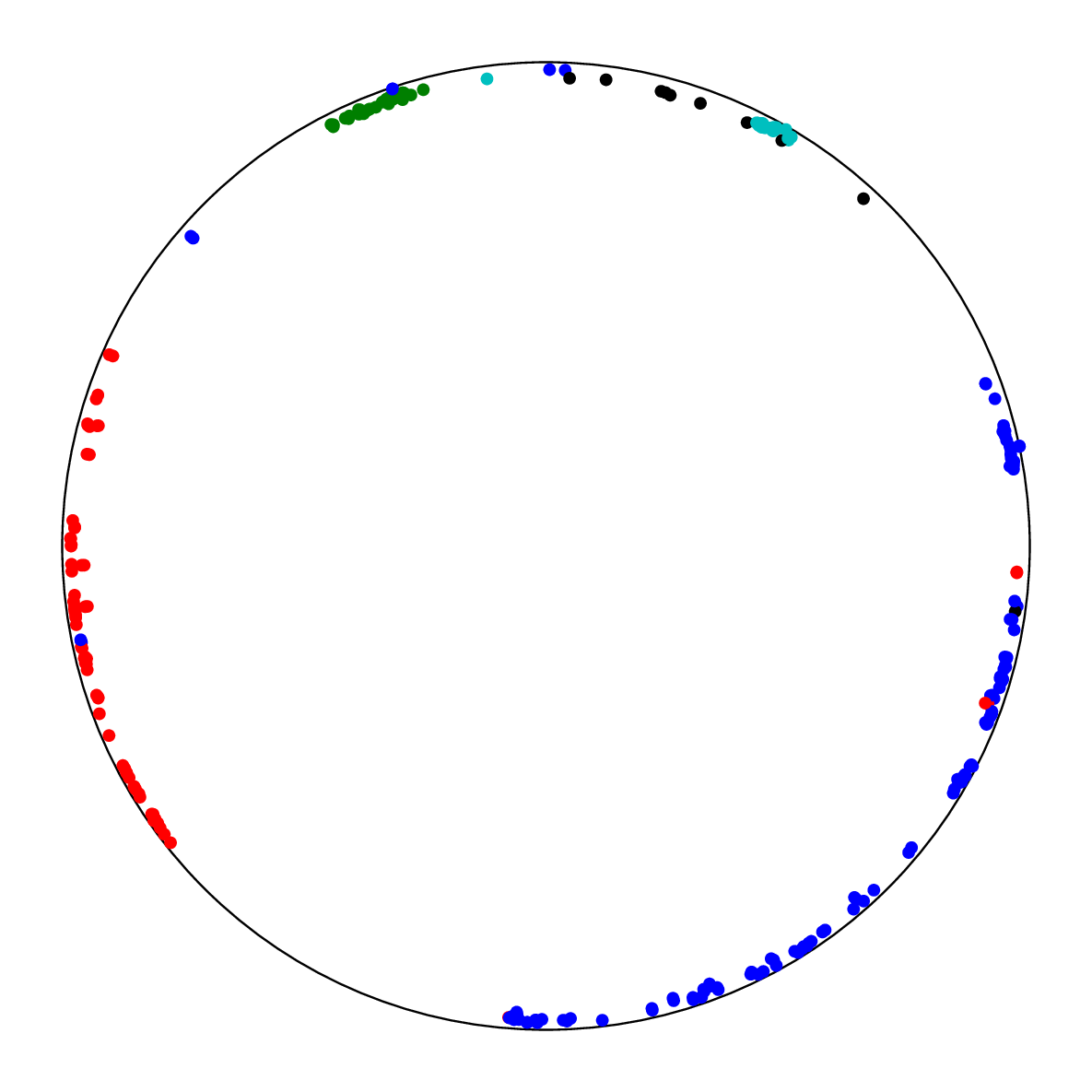}}%
    \qquad
    \subfigure[Wisconsin ($\sK$)]{\label{wisconsin_klein}%
      \includegraphics[width=0.21\linewidth]{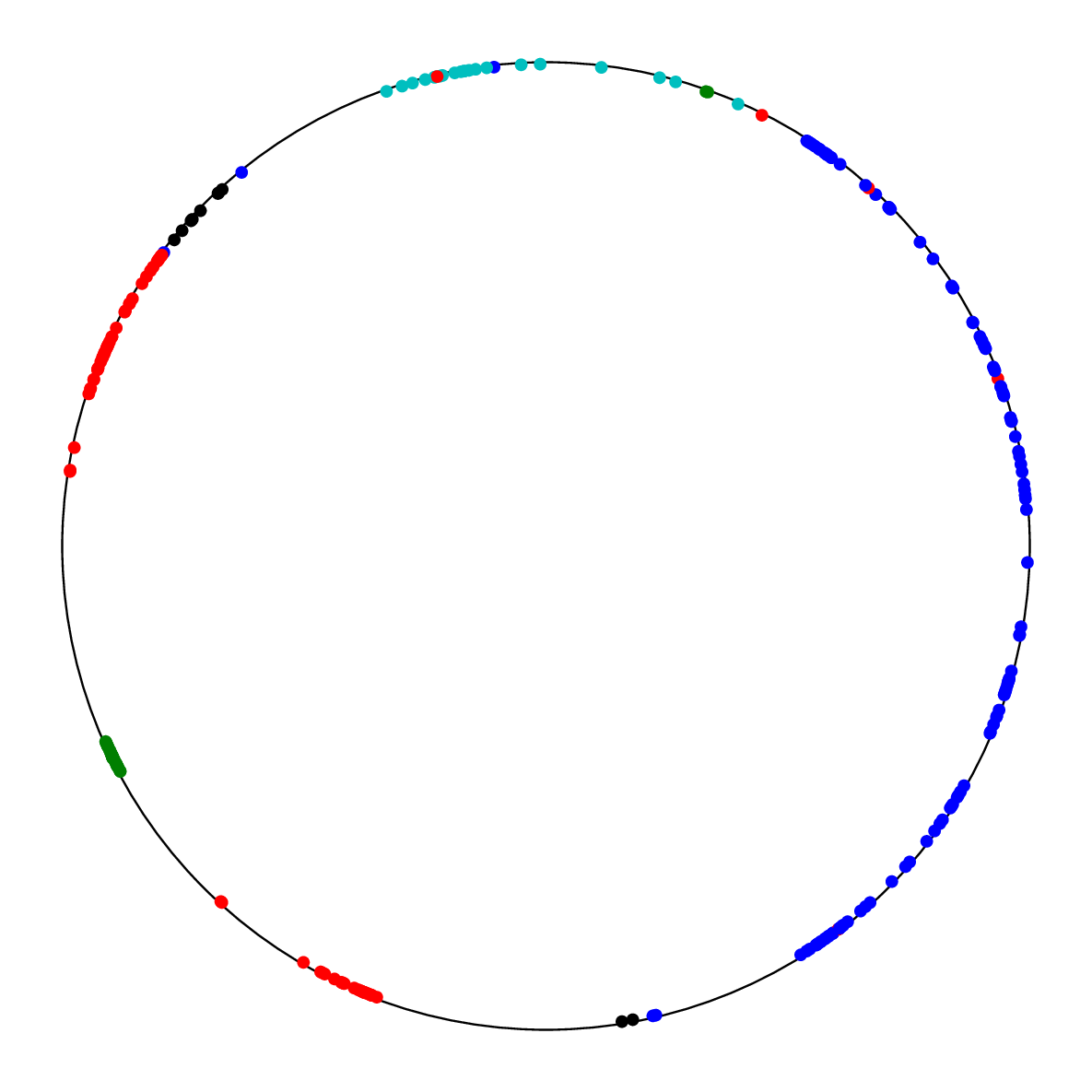}}
  }
\end{figure}

\section{Conclusion}
In this paper, we provide a detailed framework for Klein HNNs, addressing the gaps largely overlooked from previous works. We derive compact formulas for key operations in the Klein model. Subsequently, we connect the geometry of the Klein model with the Einstein gyrovector spaces, providing a simple and elegant formulation for the tangent space construction of the Klein linear layer. Our experiments show that Klein HNNs achieve comparable performance with Poincar\'e/hyperboloid HNNs while maintaining training efficiency. Furthermore, our framework supports additional operations, such as Einstein midpoints, all within the Klein model. Future works include extending the current operations to other common neural operations.


\bibliography{pmlr-sample}

\newpage
\appendix

\section{Hyperbolic Geometry}\label{apd:first}
We review basic facts about the Poincar\'e ball model and the hyperboloid model of hyperbolic space. For readers seeking a more introductory overview of hyperbolic geometry, we recommend \citet{anderson2006hyperbolic}. We also recommend the work of \citet{peng2021hyperbolic} for a survey of HNNs. For a comprehensive study of operations within these models, including those used in hyperbolic neural networks, we refer to the detailed work of \citet{ganea2018hyperbolic} and \citet{chami2019hyperbolic}. 

\subsection{The Poincar\'e Ball Model}
The $n$-dimensional Poincaré ball model with constant negative curvature of $-1$ is the Riemannian manifold $\sB^n = \left\{\rvx \in \mathbb{R}^n \mid \|\rvx\|<1\right\}$ with the metric tensor ${g}^{\sB}_\rvx=\left(\rho_{\rvx}\right)^2 \rmI_n$, where $\rho_\rvx=2\left(1-\|\rvx\|^2\right)^{-1}$ is the conformal factor and $\rmI_n$ is the Euclidean metric. The geodesic distance between two point $\rvx, \rvy \in \sB$ is given by
\begin{equation}
d_\gB(\rvx, \rvy) = \cosh^{-1}\left(1+\frac{2\|\rvx-\rvy\|^2}{\left(1-\|\rvx\|^2\right)\left(1-\|\rvy\|^2\right)}\right).
\end{equation}

Given $\rvx \in \sB^n$, $\gT_\rvx \sB^n$ denotes the tangent space of $\sB^n$ at $\rvx$. For $\rvx, \rvy \in \sB^n$ with $\rvx \neq \rvy$ and $\rvv \in \gT_\rvx \sB^n\backslash\{\mathbf{0}\}$, the exponential map $\exp^\gB_\rvx: \gT_{\rvx}\sB^n \to \sB^n$ and the logarithmic map $\log^\gB_\rvx: \sB^n \to \gT_{\rvx}\sB^n$ satisfy 
\begin{align}
\exp^\gB_\rvx(\rvv) &= \rvx \oplus_\text{M}\left(\tanh \left(\frac{\rho_\rvx\|\rvv\|}{2}\right) \frac{\rvv}{\|\rvv\|}\right), \\
\log^\gB_\rvx(\rvy) &= \frac{2}{\rho_\rvx} \tanh ^{-1}\left(\left\|-\rvx \oplus_\text{M} \rvy\right\|\right) \frac{-\rvx \oplus_\text{M} \rvy}{\left\|-\rvx \oplus_\text{M} \rvy\right\|},
\end{align}
where $\oplus_\text{M}$ is the M\"obius addition, defined by 
\begin{equation}\label{eq:mobius_add}
\rvx \oplus_{\text{M}} \rvy=\frac{\left(1+2 \rvx\cdot \rvy+\|\rvy|^2\right) \rvx+\left(1-\|\rvx\|^2\right) \rvy}{1+2 \rvx \cdot \rvy+\|\rvx\|^2\|\rvy\|^2}.
\end{equation}
In particular, the exponential and logarithmic maps at the hyperbolic origin $\rvo$ are given by
\begin{equation}
\exp^\gB_\rvo(\rvv) = \tanh(\|\rvv\|)\frac{\rvv}{\|\rvv\|}, 
\quad 
\log^\gB_\rvo(\rvy) = \tanh^{-1}(\|\rvy\|)\frac{\rvy}{\|\rvy\|}.
\end{equation}

\citet{ganea2018hyperbolic} proved that, the parallel transport w.r.t. the Levi-Civita connection of $\rvv \in \gT_{\rvo}\sB^n$ to the tangent space $\gT_{\rvx}\sB^n$ at $\rvx$ is given by 
\begin{equation}\label{eq:ganea_pt_B}
P_{\rvo\to\rvx}(\rvv) = \log^\gB_{\rvx}\left(\rvx\oplus_{\text{M}}\exp^\gB_{\rvo}(\rvv)\right).
\end{equation}

\subsection{The Hyperboloid Model}
The $n$-dimensional hyperboloid model with constant negative curvature $-1$ is represented as $\sL^{n} =\{\rvx\in\sR^{n+1}| \langle\rvx,\rvx\rangle_{\gL}=-1,x_t>0\}$, where $\langle \cdot, \cdot\rangle_{\gL}$ denotes the Minkowski inner product, $\langle\rvx,\rvy\rangle_{\gL}:= -x_0 y_0 + x_1 y_1+\cdots+x_ny_n$. It has the metric tensor $g^\sL = \operatorname{diag} \left( [-1, \vone_n^\T] \right)$. From a special relativity perspective, we may also write $\rvx\in\sL^n$ as $[x_t, \rvx_s^\T]^\T$, for $x_t$ being the time axis and $\rvx_s$ being the spatial axes. 

The tangent space of $\sL^n$ at $\rvx$ is the orthogonal space of $\sL^{n}$ with respect to the Minkowski inner product and is represented as $\gT_{\rvx}\sL^{n}:=\{\rvv \in\sR^{n+1}:\langle\rvv, \rvx \rangle_{\gL}=0\}$. Specifically, let $\{\rve_0, \cdots, \rve_n\}$ be the canonical basis of $\R^{n+1}$, the tangent space is represented with the understanding that
\begin{equation}
    \frac{\partial}{\partial x_i} \bigg\vert_\rvx = \frac{\rvx \cdot \rve_i}{\sqrt{1+\norm{\rvx}^2}} \rve_0 + \rve_i.
\end{equation}

The unit-speed geodesic $\phi_{\rvx,\rvv}(t)$ with $\phi_{\rvx,\rvv}(0) = \rvx \in \sL^n$ and $\dot{\phi}_{\rvx,\rvv}(0) = \rvv \in \gT_\rvx \sL^n$ is given by
\begin{equation}\label{eq:geodesic_L}
\phi_{\rvx,\rvv}(t) = \rvx\cosh(t) + \rvv\sinh(t).
\end{equation}
The geodesic distance of $\rvx, \rvy \in \sL^{n}$ is given by $d_{\gL}(\rvx,\rvy)=\mathrm{cosh}^{-1}(-\langle\rvx,\rvy\rangle_{\gL})$. For $\rvx, \rvy \in \sL^{n}$ with $\rvx \neq \rvy$ and $\rvv \in \gT_{\rvx}\sL^{n}\backslash\{\mathbf{0\}}$, the exponential map $\exp^\gL_\rvx: \gT_{\rvx}\sL^n \to \sL^n$ and the logarithmic map $\log^\gL_\rvx: \sL^n \to \gT_{\rvx}\sL^n$ are given by 
\begin{align}
\exp^\gL_{\rvx}(\rvv) &= \cosh\left(\|\rvv\|_{\gL}\right)\rvx+\sinh\left(\|\rvv\|_{\gL}\right)\frac{\rvv}{\|\rvv\|_{\gL}}, \\    
\log^\gL_{\rvx}(\rvy) &= d_{\gL}(\rvx,\rvy)\frac{\rvy+\langle\rvx,\rvy\rangle_{\gL}\rvx}{\|\rvy+\langle\rvx,\rvy\rangle_{\gL}\rvx\|_{\gL}},
\end{align} 
where $\|\rvv\|_\gL = \sqrt{\langle\rvv,\rvv\rangle_{\gL}}$ is the induced norm from the Minkowski inner product.

The parallel transport of a tangent vector $\rvv \in \gT_{\rvo}\sL^n$ at the hyperbolic origin $\rvo = [1, \vzero^\T]^\T$ to the tangent space $\gT_{\rvx^\gL}\sL^n$ is given by
\begin{equation}\label{eq:parallel_trans_L}
P_{\rvo\to\rvx}(\rvv) = \rvv-\frac{\langle\log_{\rvo}(\rvx),\rvv\rangle_{\gL}}{d_{\gL}(\rvo,\rvx)^2}(\log_{\rvo}(\rvx)+\log_{\rvx}(\rvo)).
\end{equation}

\section{Additional Proofs}\label{apd:second}
\subsection{Proof of \lemmaref{lem:tangent vectors in Klein and Poincare}}
1. The corresponding tangent vectors $\rvv^\gB \in \gT_{\rvx^\gB}\sB^n$ and $\rvv^\gK \in \gT_{\rvx^\gK}\sK^n$ are related using Jacobian matrices. Namely,
\begin{equation}
\rvv^\gK = \frac{\partial \pi_{\gB\to\gK}(\rvx^\gB)}{\partial \rvx^\gB} \rvv^\gB, \quad \rvv^\gB = \frac{\partial \pi_{\gK\to\gB}(\rvx^\gK)}{\partial \rvx^\gK} \rvv^\gK. 
\end{equation}
Since
\begin{equation}
\frac{\partial \pi_{\gB\to\gK}(\rvx^\gB)}{\partial \rvx^\gB} = \frac{2}{1+\|\rvx^\gB\|^2} \rmI + \rvx^\gB \cdot \frac{-2\cdot2({\rvx^\gB})^\T}{(1+\|\rvx^\gB\|^2)^2},
\end{equation}
\begin{equation}
\frac{\partial \pi_{\gK\to\gB}(\rvx^\gK)}{\partial \rvx^\gK} = \frac{1}{1+\sqrt{1-\|\rvx^\gK\|^2}} \rmI + \rvx^\gK \cdot \frac{-1\cdot2({\rvx^\gK})^\T}{2\sqrt{1-\|\rvx^\gK\|^2}(1+\sqrt{1-\|\rvx^\gK\|^2})^2},
\end{equation}
we have
\begin{equation}
\begin{aligned}
\rvv^\gK = \frac{\partial \pi_{\gB\to\gK}(\rvx^\gB)}{\partial \rvx^\gB} \rvv^\gB = \frac{2}{1+\|\rvx^\gB\|^2} \rvv^\gB - \frac{4\rvx^\gB\cdot\rvv^\gB}{(1+\|\rvx^\gB\|^2)^2}\rvx^\gB,
\end{aligned}
\end{equation}
\begin{equation}
\rvv^\gK = \frac{\partial \pi_{\gK\to\gB}(\rvx^\gK)}{\partial \rvx^\gK} \rvv^\gK = \frac{1}{1+\sqrt{1-\|\rvx^\gK\|^2}} \rvv^\gK +\frac{\rvx^\gK\cdot\rvv^\gK}{\sqrt{1-\|\rvx^\gK\|^2}(1+\sqrt{1-\|\rvx^\gK\|^2})^2}  \rvx^\gK.
\end{equation}

\noindent 2. Similarly,
\begin{equation}
\rvv^\gK = \frac{\partial \pi_{\gL\to\gK}(\rvx^\gL)}{\partial \rvx^\gL} \rvv^\gL, \quad \rvv^\gL = \frac{\partial \pi_{\gK\to\gL}(\rvx^\gK)}{\partial \rvx^\gK} \rvv^\gK. \end{equation}

For $\frac{\partial \pi_{\gL\to\gK}(\rvx^\gL)}{\partial \rvx^\gL}$, let $\rvy = \pi_{\gL\to\gK}(\rvx^\gL)$. Then 
\begin{equation}
\frac{\partial \pi_{\gL\to\gK}(\rvx^\gL)}{\partial \rvx^\gL} = \frac{\partial \rvy}{\partial \rvx^\gL} 
= \begin{bmatrix}\frac{\partial \rvy}{\partial x_t}&\frac{\partial \rvy}{\partial \rvx_s}\end{bmatrix},
\end{equation}
with $\frac{\partial \rvy}{\partial x_t}$ being a $n\times 1$ column vector and $\frac{\partial \rvy}{\partial \rvx_s}$ being a $n\times n$ matrix. 
The isometric mapping in~\equationref{eq:iso_map_L_K} indicates that $\rvy = \frac{1}{x_t}\rvx_s$, and thus 
\begin{equation}
\frac{\partial \rvy}{\partial x_t} = -\frac{1}{x_t^2}\rvx_s, \quad \frac{\partial \rvy}{\partial \rvx_s} = \frac{1}{x_t}\rmI.
\end{equation}
Writing $\rvv^\gL$ as $\left[v_t, \rvv_s^\T\right]^\T$, we have
\begin{equation}
\begin{aligned}
\rvv^\gK &= \frac{\partial \pi_{\gL\to\gK}(\rvx^\gL)}{\partial \rvx^\gL} \rvv^\gL = \begin{bmatrix}\frac{\partial \rvy}{\partial x_t}&\frac{\partial \rvy}{\partial \rvx_s}\end{bmatrix}\begin{bmatrix}v_t\\\rvv_s\end{bmatrix}= v_t \frac{\partial \rvy}{\partial x_t} + \frac{\partial \rvy}{\partial \rvx_s}\rvv_s = -\frac{v_t}{x_t^2}\rvx_s + \frac{1}{x_t}\rvv_s.
\end{aligned}
\end{equation}
Hence, 
\begin{equation}
\rvv^\gK = \frac{\partial \pi_{\gL\to\gK}(\rvx^\gL)}{\partial \rvx^\gL} \rvv^\gL = -\frac{v_t}{x_t^2}\rvx_s + \frac{1}{x_t}\rvv_s \in \gT_{\rvx^\gK}\sK^n.
\end{equation}

For $\frac{\partial \pi_{\gK\to\gL}(\rvx^\gK)}{\partial \rvx^\gK}$, let $\rvz = \pi_{\gK\to\gL}(\rvx^\gK) =\left[z_t, \rvz_s^\T\right]^\T$. Then
\begin{equation}
\frac{\partial \pi_{\gK\to\gL}(\rvx^\gK)}{\partial \rvx^\gK} = \frac{\partial \rvz}{\partial \rvx^\gK}  
=\begin{bmatrix}\frac{\partial z_t}{\partial \rvx^\gK}\\\frac{\partial \rvz_s}{\partial \rvx^\gK}\end{bmatrix}.
\end{equation}
with $\frac{\partial z_t}{\partial \rvx^\gK}$ being a $1\times n$ row vector and $\frac{\partial \rvz_s}{\partial \rvx^\gK}$ being a $n\times n$ matrix. From the mapping between the hyperboloid model and the Klein model, we know
\begin{equation}
z_t = \frac{1}{\sqrt{1-\|\rvx^\gK\|^2}}, \quad \rvz_s = \frac{1}{\sqrt{1-\|\rvx^\gK\|^2}}\rvx^\gK.
\end{equation}
We compute $\frac{\partial z_t}{\partial \rvx^\gK}$ and $\frac{\partial \rvz_s}{\partial \rvx^\gK}$ respectively: 
\begin{align}
\frac{\partial z_t}{\partial \rvx^\gK} &= -\frac{1}{2}\left(1-\|\rvx^\gK\|^2\right)^{-\frac{3}{2}}\cdot\left(-2(\rvx^\gK)^\T\right) = \frac{(\rvx^\gK)^\T}{\left(1-\|\rvx^\gK\|^2\right)^{\frac{3}{2}}};\\
\frac{\partial \rvz_s}{\partial \rvx^\gK} &= \frac{1}{\sqrt{1-\|\rvx^\gK\|^2}}\rmI + \rvx^\gK \cdot \frac{(\rvx^\gK)^\T}{\left(1-\|\rvx^\gK\|^2\right)^{\frac{3}{2}}}.
\end{align}
Putting everything together, 
\begin{equation}
\begin{aligned}
\rvv^\gL = \frac{\partial \pi_{\gK\to\gL}(\rvx^\gK)}{\partial \rvx^\gK} \rvv^\gK = \begin{bmatrix}\frac{\rvx^\gK\cdot\rvv^\gK}{\left(1-\|\rvx^\gK\|^2\right)^{\frac{3}{2}}}\\\frac{1}{\sqrt{1-\|\rvx^\gK\|^2}}\rvv^\gK + \frac{\rvx^\gK\cdot \rvv^\gK }{\left(1-\|\rvx^\gK\|^2\right)^{\frac{3}{2}}} \rvx^\gK \end{bmatrix} \in \gT_{\rvx^\gL}\sL^n.
\end{aligned}
\end{equation}
\qed

\subsection{Proof of \lemmaref{lem:geodesic}}
The proof is analogous to that of $\sB^n$~\citep[Theorem 1]{ganea2018hyperbolic2}.
Let $\rvx^\gL \in \sL^n$ and $\rvv^\gL \in \gT_{\rvx^\gL} \sL^n$ with $\langle \rvv^\gL, \rvv^\gL \rangle_{\gL} = 1$. According to \equationref{eq:geodesic_L}, the unit-speed geodesic $\phi(t)=\phi_{\rvx^\gL,\rvv^\gL}(t)$ with $\phi_{\rvx^\gL,\rvv^\gL}(0) = \rvx^\gL$ and $\dot{\phi}_{\rvx^\gL,\rvv^\gL}(0) = \rvv^\gL$ is 
\begin{equation}\label{eq:geodesic_L_lem_geodesic}
\phi_{\rvx^\gL,\rvv^\gL}(t) = \rvx^\gL \cosh(t) + \rvv^\gL \sinh(t).
\end{equation}
Let $\rvx^\gK \in \sK^n$ and $\rvv^\gK \in \gT_\rvx^\gK \sK^n$ with $g_{\rvx^{\gK}}(\rvv^\gK, \rvv^\gK) = 1$. Let $\gamma(t)=\gamma_{\rvx^\gK,\rvv^\gK}(t) \in \sK^n$ denote the unit-speed geodesic in $\sK^n$ with $\gamma_{\rvx^\gK,\rvv^\gK}(0) = \rvx^\gK$ and $\dot{\gamma}_{\rvx^\gK,\rvv^\gK}(0) = \rvv^\gK$.
Suppose $\rvx^\gL$ and $\rvx^\gK$ are corresponding points in $\sL^n$ and $\rvv^\gL$ and $\rvv^\gK$ are corresponding vectors in their tangent spaces. Then
\begin{align}
& \rvx^\gL = \pi_{\gK \rightarrow \gL} (\rvx^\gK) = \frac{1}{\sqrt{1-\|\rvx^\gK\|^2}} 
\begin{bmatrix}
1 \\
\rvx^\gK
\end{bmatrix}
= \lambda_{\rvx^\gK}
\begin{bmatrix}
1 \\
\rvx^\gK
\end{bmatrix}
,
\\
& \rvv^\gL = \dot{\phi}(0) = \left. \frac{\partial \pi_{\gK \rightarrow \gL} (\rvy^\gK)}{\partial \rvy^\gK} \right\vert_{\gamma(0)} \dot{\gamma}(0) = \frac{\partial \pi_{\gK \rightarrow \gL} (\rvx^\gK)}{\partial \rvx^\gK} \rvv^\gK.
\end{align}
With
\begin{equation}
\frac{\partial \lambda_{\rvx^\gK}}{\partial \rvx^\gK} = \frac{\rvx^\gK}{(1-\| \rvx^\gK \|^2)^{3/2}} = \lambda^3_{\rvx^\gK} \rvx^\gK,
\end{equation}
it follows that
\begin{equation}
\begin{aligned}
\rvv^\gL &= \frac{\partial \pi_{\gK \rightarrow \gL} (\rvx^\gK)}{\partial \rvx^\gK} \rvv^\gK = \left( \frac{\partial}{\partial \rvx^\gK} \left(
\begin{bmatrix}
\lambda_{\rvx^\gK} \\
\lambda_{\rvx^\gK} \rvx^\gK
\end{bmatrix}
\right) \right) \rvv^\gK = \begin{bmatrix}
\frac{\partial \lambda_{\rvx^\gK}}{\partial \rvx^\gK} \\
\frac{\partial \lambda_{\rvx^\gK}}{\partial \rvx^\gK}(\rvx^\gK)^\T + \lambda_{\rvx^\gK} {\bf I}
\end{bmatrix}
\rvv^\gK
\\
&= \begin{bmatrix}
\lambda^3_{\rvx^\gK} \left(\rvx^\gK\cdot\rvv^\gK\right) \\
\lambda^3_{\rvx^\gK} \left(\rvx^\gK\cdot\rvv^\gK\right) \rvx^\gK  + \lambda_{\rvx^\gK} \rvv^\gK
\end{bmatrix}.
\end{aligned}
\end{equation}
Making substitutions in \equationref{eq:geodesic_L_lem_geodesic} yields
\begin{equation}
\phi(t) = \lambda_{\rvx^\gK}
\begin{bmatrix}
1 \\
\rvx^\gK
\end{bmatrix}
\cosh(t) + 
\begin{bmatrix}
\lambda^3_{\rvx^\gK} \left(\rvx^\gK\cdot\rvv^\gK\right) \\
\lambda^3_{\rvx^\gK} \left(\rvx^\gK\cdot\rvv^\gK\right) \rvx^\gK  + \lambda_{\rvx^\gK} \rvv^\gK
\end{bmatrix}
\sinh(t) .
\end{equation}
Since the geodesics are preserved by the isometric mapping, $\gamma$ is given by
\begin{equation}
\begin{aligned}
\gamma_{\rvx^\gK,\rvv^\gK}(t) &= \pi_{\gL \rightarrow \gK} \circ \phi(t) \\
&= \frac{\left(\lambda_{\rvx^\gK} \cosh(t) + \lambda^3_{\rvx^\gK} \left(\rvx^\gK\cdot\rvv^\gK \right) \sinh(t)\right) \rvx^\gK + \left(\lambda_{\rvx^\gK} \sinh(t)\right) \rvv^\gK} {\lambda_{\rvx^\gK} \cosh(t) + \lambda^3_{\rvx^\gK} \left(\rvx^\gK\cdot\rvv^\gK \right)\sinh(t)} \\
&= \frac{\left( \cosh(t) + \lambda^2_{\rvx^\gK} \left(\rvx^\gK\cdot\rvv^\gK \right) \sinh(t)\right) \rvx^\gK + \sinh(t) \rvv^\gK} {\cosh(t) + \lambda^2_{\rvx^\gK} \left(\rvx^\gK\cdot\rvv^\gK \right)\sinh(t)} \\
&= \rvx^\gK + \frac{\sinh(t) \rvv^\gK} {\cosh(t) + \lambda^2_{\rvx^\gK} \left(\rvx^\gK\cdot\rvv^\gK \right)\sinh(t)} .
\end{aligned}
\end{equation}
\qed

\subsection{Proof of Proposition \ref{pro:parallel transport in Klein}}

Write $\rvx^{\gL}$ as $\left[x_t, \rvx_s^\T\right]^\T$. Since $\rvo^{\gL}=[1, 0, \cdots, 0]^\T$, we have $d_{\gL}(\rvo^{\gL},\rvx^{\gL}) = \mathrm{cosh^{-1}}(x_t)$. Then
\begin{equation}
\begin{aligned}
\log_{\rvo^{\gL}}(\rvx^{\gL})
&=d_{\gL}(\rvo^{\gL},\rvx^{\gL})\frac{\rvx^{\gL}+\langle\rvo^{\gL},\rvx^{\gL}\rangle_{\gL}\rvo^{\gL}}{\|\rvx^{\gL}+\langle\rvo^{\gL},\rvx^{\gL}\rangle_{\gL}\rvo^{\gL}\|_{\gL}}=\mathrm{cosh^{-1}}(x_t)\frac{\rvx^{\gL}-x_t\rvo^{\gL}}{\|\rvx^{\gL}-x_t\rvo^{\gL}\|_{\gL}}.
\end{aligned}
\end{equation}
Here, $\rvx^{\gL} - x_t\rvo^{\gL} = \left[x_t, \rvx_s^\T\right]^\T - \left[x_t, \mathbf{0}^\T\right]^\T = \left[0, \rvx_s^\T\right]^\T $ and thus $\|\rvx^{\gL}-x_t\rvo^{\gL}\|_{\gL} = \|\rvx_s\|$.
Therefore, $\log_{\rvo^{\gL}}(\rvx^{\gL}) = \frac{\mathrm{cosh^{-1}}(x_t)}{\|\rvx_s\|}\left[0, \rvx_s^\T\right]^\T$. Similarly, 
\begin{equation}
\begin{aligned}
\log_{\rvx^{\gL}}(\rvo^{\gL})
&=d_{\gL}(\rvx^{\gL},\rvo^{\gL})\frac{\rvo^{\gL}+\langle\rvx^{\gL},\rvo^{\gL}\rangle_{\gL}\rvx^{\gL}}{\|\rvo^{\gL}+\langle\rvx^{\gL},\rvo^{\gL}\rangle_{\gL}\rvx^{\gL}\|_{\gL}}=\mathrm{cosh^{-1}}(x_t)\frac{\rvo^{\gL}-x_t\rvx^{\gL}}{\|\rvo^{\gL}-x_t\rvx^{\gL}\|_{\gL}}.
\end{aligned}
\end{equation}
Here, $\rvo^{\gL}-x_t\rvx^{\gL} = \left[1, \mathbf{0}^\T\right]^\T - x_t \left[x_t, \rvx_s^\T \right]^\T = \left[1-x_t^2, x_t\rvx_s^\T\right]^\T$, and thus
\begin{equation}
\begin{aligned}
\|\rvo^{\gL}-x_t\rvx^{\gL}\|_\gL
&=\sqrt{-\left(1-x_t^2\right)^2+x_t^2\|\rvx_s\|^2}=\sqrt{-\left(-\|\rvx_s\|^2\right)^2+ x_t^2\|\rvx_s\|^2}\\
&=\sqrt{\|\rvx_s\|^2(x_t^2-\|\rvx_s\|^2)}=\sqrt{\|\rvx_s\|^2}=\|\rvx_s\|.
\end{aligned}
\end{equation}
Therefore, $\log_{\rvx^{\gL}}(\rvo^{\gL}) = \frac{\mathrm{cosh^{-1}}(x_t)}{\|\rvx_s\|}\left[1-x_t^2, x_t\rvx_s^\T\right]^\T$. Since $\rvv^{\gL} \in \gT_{\rvo^\gL}\sL^n$, it holds that $\langle\rvo, \rvv^{\gL}\rangle_\gL = 0$. Writing $\rvv^{\gL}$ as $\left[v_t, \rvv_s^\T\right]^\T$, this indicates that $v_t=0$. Hence, 
\begin{equation}
\begin{aligned}
P_{\rvo^{\gL}\to\rvx^{\gL}}^{\gL}(\rvv^{\gL}) &= \rvv^{\gL}-\frac{\langle\log_{\rvo}^{\gL}(\rvx^{\gL}),\rvv^{\gL}\rangle_{\gL}}{d_{\gL}(\rvo^{\gL},\rvx^{\gL})^2}(\log_{\rvo^{\gL}}^{\gL}(\rvx^{\gL})+\log_{\rvx^{\gL}}^{\gL}(\rvo^{\gL})) \\
&= \begin{bmatrix}0\\\rvv_s\end{bmatrix} - \frac{\frac{\mathrm{cosh^{-1}}(x_t)}{\|\rvx_s\|}\rvx_s\cdot\rvv_s}{\left(\mathrm{cosh^{-1}}(x_t)\right)^2}\frac{\mathrm{cosh^{-1}}(x_t)}{\|\rvx_s\|}\left(\begin{bmatrix}0\\\rvx_s\end{bmatrix}+\begin{bmatrix}1-x_t^2\\x_t\rvx_s\end{bmatrix}\right)\\
&= \begin{bmatrix}0\\\rvv_s\end{bmatrix} - \frac{\rvx_s\cdot\rvv_s}{\|\rvx_s\|^2}\begin{bmatrix}1-x_t^2\\(1+x_t)\rvx_s\end{bmatrix}=
\begin{bmatrix}\rvx_s\cdot\rvv_s\\\rvv_s-\frac{\rvx_s\cdot\rvv_s}{(x_t-1)}\rvx_s \end{bmatrix}.
\end{aligned}
\end{equation}
Next, we use the isometric mapping in \equationref{eq:iso_map_L_K} as well as Equations~\eqref{eq:yidan_review_1} and \eqref{eq:yidan_review_2} to derive $P_{\rvo^{\gK}\to\rvx^{\gK}}(\rvv^{\gK})$. 
For clarity, denote $P_{\rvo^{\gL}\to\rvx^{\gL}}(\rvv^{\gL}) = \rvw^\gL = \left[w_t, \rvw_s^\T\right]^\T \in \gT_{\rvx^\gL}\sL^n$. Then, by \lemmaref{lem:tangent vectors in Klein and Poincare}, the corresponding tangent vector in $\gT_{\rvx^\gK}\sK^n$ is given by 
\begin{equation}
\begin{aligned}
\rvw^\gK &= -\frac{w_t}{x_t^2}\rvx_s + \frac{1}{x_t}\rvw_s= -\frac{\rvx_s\cdot\rvv_s}{x_t^2}\rvx_s+\frac{1}{x_t}\left(\rvv_s-\frac{\rvx_s\cdot\rvv_s}{(x_t-1)}\rvx_s\right).
\end{aligned}
\end{equation}
Expressing $x_t, \rvx_s$ in terms of $\rvx^\gK$, we have 
\begin{equation}
\begin{aligned}
x_t = \frac{1}{\sqrt{1-\|\rvx^\gK\|^2}}, \quad \rvx_s = \frac{1}{\sqrt{1-\|\rvx^\gK\|^2}}\rvx^\gK.
\end{aligned}
\end{equation}
We further express $\rvv^\gL\in\gT_{\rvo^\gL}\sL^n$ in terms of the corresponding tangent vector $\rvv^\gK$ in $\gT_{\rvo^\gK}\sK^n$. We already know that $v_t =0$. Also, 
\begin{equation}
\rvv_s = \frac{1}{\sqrt{1-\|\rvo^\gK\|^2}}\rvv^\gK + \frac{\rvo^\gK\cdot \rvv^\gK }{\left(1-\|\rvo^\gK\|^2\right)^{\frac{3}{2}}} \rvo^\gK = \rvv^\gK.
\end{equation}
Hence, 
\begin{align}
P_{\rvo^{\gL}\to\rvx^{\gL}}(\rvv^{\gL}) = \rvw^\gK &= -\frac{\rvx_s\cdot\rvv_s}{x_t^2}\rvx_s+\frac{1}{x_t}\left(\rvv_s-\frac{\rvx_s\cdot\rvv_s}{(x_t-1)}\rvx_s\right) \nonumber \\
&= -\left(1-\|\rvx^\gK\|^2\right) \frac{\rvx^\gK\cdot\rvv^\gK}{\sqrt{1-\|\rvx^\gK\|^2}}\frac{1}{\sqrt{1-\|\rvx^\gK\|^2}}\rvx^\gK \nonumber  \\
&\quad + \sqrt{1-\|\rvx^\gK\|^2}\left(\rvv^\gK-\frac{1}{\frac{1}{\sqrt{1-\|\rvx^\gK\|^2}}-1}\frac{\rvx^\gK\cdot\rvv^\gK}{\sqrt{1-\|\rvx^\gK\|^2}}\frac{1}{\sqrt{1-\|\rvx^\gK\|^2}}\rvx^\gK\right) \nonumber \\
&= -(\rvx^\gK\cdot\rvv^\gK)\rvx^\gK+\sqrt{1-\|\rvx^\gK\|^2} \rvv^\gK - \frac{\rvx^\gK\cdot\rvv^\gK}{1-\sqrt{1-\|\rvx^\gK\|^2}}\rvx^\gK \nonumber \\
&= \frac{(\rvx^\gK\cdot\rvv^\gK)(\sqrt{1-\|\rvx^\gK\|^2}-2)}{1-\sqrt{1-\|\rvx^\gK\|^2}}\rvx^\gK +\sqrt{1-\|\rvx^\gK\|^2} \rvv^\gK.
\end{align}
\qed


\subsection{Proof of \theoremref{thm:Klein and Einstein scalar multiplication}}

Recall that \corollaryref{cor:exp and log maps} indicates
\begin{align}
\exp^\gK_{\rvo^\gK}(\rvv^\gK) 
&= \tanh(\|\rvv^\gK\|)\frac{\rvv^\gK}{\|\rvv^\gK\|}, \\
\log^\gK_{\rvo^\gK}(\rvx^\gK) 
&= \cosh^{-1} \left(\frac{1}{\sqrt{1-\|\rvx^\gK\|^2}}\right) \frac{\rvx^\gK}{\| \rvx^\gK\|}.
\end{align}
On the one hand,
\begin{equation}
\begin{aligned}
\exp^\gK_{\rvo^\gK}(r\log^\gK_{\rvo^\gK}( \rvx^\gK)) 
&=\exp^\gK_{\rvo^\gK}\left(r\cosh^{-1} \left(\frac{1}{\sqrt{1-\|\rvx^\gK\|^2}}\right) \frac{\rvx^\gK}{\| \rvx^\gK\|}\right)\\
&=\tanh\left(|r|\cosh^{-1} \left(\frac{1}{\sqrt{1-\|\rvx^\gK\|^2}}\right)\right)\frac{r\rvx^\gK}{|r|\|\rvx^\gK\|}\\
&=\tanh\left(r\cosh^{-1} \left(\frac{1}{\sqrt{1-\|\rvx^\gK\|^2}}\right) \right)\frac{\rvx^\gK}{\|\rvx^\gK\|}.
\end{aligned}
\end{equation}
On the other hand, 
\begin{equation}
r\otimes_\text{E} \rvx^\gK = \tanh{\left(r\tanh^{-1}\left(\|\rvx^\gK\|\right)\right)}\frac{\rvx^\gK}{\|\rvx^\gK\|}.
\end{equation}
Therefore, to prove $r\otimes_\text{E} \rvx^\gK = \exp^\gK_{\rvo^\gK}\left(r\log^\gK_{\rvo^\gK}\left(\rvx^\gK\right)\right)$, it suffices to show: 
\begin{equation}
\cosh^{-1} \left(\frac{1}{\sqrt{1-\|\rvx^\gK\|^2}}\right) = \tanh^{-1}\left(\|\rvx^\gK\|\right).
\end{equation}
Let $\cosh^{-1} \left(\frac{1}{\sqrt{1-\|\rvx^\gK\|^2}}\right) = t$, then $\cosh(t)=\frac{1}{\sqrt{1-\|\rvx^\gK\|^2}}$, i.e.,
\begin{equation}
\frac{e^{t}+e^{-t}}{2}=\frac{1}{\sqrt{1-\|\rvx^\gK\|^2}}.
\end{equation}
This implies that
\begin{equation}
\begin{aligned}
\|\rvx^\gK\|^2 &= 1-\frac{4}{e^{2t}+e^{-2t}+2}= \frac{e^{2t}+e^{-2t}-2}{e^{2t}+e^{-2t}+2} = \left(\frac{e^{t}-e^{-t}}{e^{t}+e^{-t}}\right)^2.
\end{aligned}
\end{equation}
Therefore, $\|\rvx^\gK\| = \frac{e^{t}-e^{-t}}{e^{t}+e^{-t}}$ since $t>0$, i.e. $\tanh^{-1}\left(\|\rvx^\gK\|\right) = t$, completing the proof.
\qed

\subsection{Proof of \theoremref{thm:Klein and Einstein addition}}
Our proof is based on \equationref{eq:ganea_pt_B} and the isometric mappings in Equations \eqref{eq:iso_map_B_K}--\eqref{eq:yidan_review_b2}.

Note that proving \equationref{eq:yidan_pt_K} 
is equivalent to showing that $P_{\rvo^{\gK}\to\rvx^{\gK}}(\rvv^{\gK})$ given by $\log^\gK_{\rvx^\gK}\left(\rvx^{\gK}\oplus_{\text{E}}\exp^\gK_{\rvo^\gK}(\rvv^{\gK})\right)$ and $P_{\rvo^{\gB}\to\rvx^{\gB}}(\rvv^{\gB})$ given by $\log^\gB_{\rvx^\gB}\left(\rvx^{\gB}\oplus_{\text{M}}\exp^\gB_{\rvo^\gB}(\rvv^{\gB})\right)$ are corresponding tangent vectors. This is further equivalent to showing that $\rvx^{\gK}\oplus_{\text{E}}\exp^\gK_{\rvo^\gK}(\rvv^{\gK})$ and $\rvx^{\gB}\oplus_{\text{M}}\exp^\gB_{\rvo^\gB}(\rvv^{\gB})$ are corresponding points in hyperbolic space.

On the one hand, 
applying \equationref{eq:mobius_add}, we have
\begin{equation}
\begin{aligned}
&\quad \rvx^{\gB}\oplus_{\text{M}}\exp^\gB_{\rvo^\gB}(\rvv^{\gB})= \rvx^{\gB}\oplus_{\text{M}}\left(\tanh{\left(\|\rvv^{\gB}\|\right)}\frac{\rvv^{\gB}}{\|\rvv^{\gB}\|}\right)\\
&= \frac{\left(1+\frac{2\tanh{\left(\|\rvv^{\gB}\|\right)}}{\|\rvv^{\gB}\|} \rvx^\gB\cdot \rvv^\gB+\tanh^2{\left(\|\rvv^{\gB}\|\right)}\right) \rvx^\gB+\left(1-\|\rvx^\gB\|^2\right) \frac{\tanh{\left(\|\rvv^{\gB}\|\right)}}{\|\rvv^{\gB}\|}\rvv^{\gB}}{1+\frac{2\tanh{\left(\|\rvv^{\gB}\|\right)}}{\|\rvv^{\gB}\|} \rvx^\gB\cdot \rvv^\gB+\tanh^2{\left(\|\rvv^{\gB}\|\right)}\|\rvx^\gB\|^2}\\
&= \frac{\left(\|\rvv^{\gB}\|+2\tanh{\left(\|\rvv^{\gB}\|\right)} \rvx^\gB\cdot \rvv^\gB+\tanh^2{\left(\|\rvv^{\gB}\|\right)}\|\rvv^{\gB}\|\right) \rvx^\gB+\left(1-\|\rvx^\gB\|^2\right) \tanh{\left(\|\rvv^{\gB}\|\right)}\rvv^{\gB}}{\|\rvv^{\gB}\|+2\tanh{\left(\|\rvv^{\gB}\|\right)} \rvx^\gB\cdot \rvv^\gB+\tanh^2{\left(\|\rvv^{\gB}\|\right)}\|\rvx^\gB\|^2\|\rvv^{\gB}\|}.
\end{aligned}  
\end{equation}
On the other hand, applying \equationref{eq:ein-add},
we have 
\begin{align*}
&\quad\rvx^{\gK}\oplus_\text{E}\exp^\gK_{\rvo^\gK}(\rvv^{\gK}) =  \rvx^{\gK}\oplus_\text{E}\left(\tanh(\|\rvv^\gK\|)\frac{\rvv^\gK}{\|\rvv^\gK\|}\right)\\
&= \frac1{1+\frac{\tanh(\|\rvv^\gK\|)}{\|\rvv^\gK\|}\rvx^\gK\cdot\rvv^\gK}   \Bigg( 
   \rvx^\gK+\sqrt{1-\|\rvx^\gK\|^2}\frac{\tanh(\|\rvv^\gK\|)}{\|\rvv^\gK\|}\rvv^\gK \\
&\qquad +\frac{\frac1{\sqrt{1-\|\rvx^\gK\|^2}}}{1+\frac1{\sqrt{1-\|\rvx^\gK\|^2}}}\frac{\tanh(\|\rvv^\gK\|)}{\|\rvv^\gK\|}(\rvx^\gK\cdot\rvv^\gK)\rvx^\gK \Bigg)\\
&= \frac{\|\rvv^\gK\|}{\|\rvv^\gK\|+\tanh(\|\rvv^\gK\|)\rvx^\gK\cdot\rvv^\gK}\Bigg(\rvx^\gK+\sqrt{1-\|\rvx^\gK\|^2}\frac{\tanh(\|\rvv^\gK\|)}{\|\rvv^\gK\|}\rvv^\gK\\
&\qquad +\frac{1}{1+\sqrt{1-\|\rvx^\gK\|^2}}\frac{\tanh(\|\rvv^\gK\|)}{\|\rvv^\gK\|}(\rvx^\gK\cdot\rvv^\gK)\rvx^\gK\Bigg)\\
&= \frac{\|\rvv^\gK\|}{\|\rvv^\gK\|+\tanh(\|\rvv^\gK\|)\rvx^\gK\cdot\rvv^\gK}\rvx^\gK+\frac{\tanh(\|\rvv^\gK\|)\rvx^\gK\cdot\rvv^\gK}{\|\rvv^\gK\|+\tanh(\|\rvv^\gK\|)\rvx^\gK\cdot\rvv^\gK}\frac{\rvx^\gK}{1+\sqrt{1-\|\rvx^\gK\|^2}}\\
&\qquad + \frac{\tanh(\|\rvv^\gK\|)\sqrt{1-\|\rvx^\gK\|^2}}{\|\rvv^\gK\|+\tanh(\|\rvv^\gK\|)\rvx^\gK\cdot\rvv^\gK}\rvv^\gK.
\end{align*}

Given $\rvx^\gB, \rvv^\gB$, the corresponding $\rvx^\gK\in \sK^n, \rvv^\gK \in \gT_{\rvo^\gK}\sK^n$ are
\begin{equation}
\begin{aligned}
\rvx^\gK = \frac{2}{1+\|\rvx^\gB\|^2}\rvx^\gB, \quad
\rvv^\gK = \frac{2}{1+\|\rvo^\gB\|^2} \rvv^\gB - \frac{4\rvo^\gB\cdot\rvv^\gB}{(1+\|\rvo^\gB\|^2)^2}\rvo^\gB = 2\rvv^\gB.
\end{aligned}
\end{equation}
Writing $\rvx^{\gK}\oplus_\text{E}\exp^\gK_{\rvo^\gK}(\rvv^{\gK})$ in terms of $\rvx^\gB, \rvv^\gB$ yields 
\begin{align*}
& \quad \rvx^{\gK}\oplus_\text{E}\exp^\gK_{\rvo^\gK}(\rvv^{\gK}) \\
&= \frac{2\|\rvv^\gB\|}{2\|\rvv^\gB\|+\tanh(2\|\rvv^\gB\|)\frac{2\rvx^\gB\cdot2\rvv^\gB}{1+\|\rvx^\gB\|^2}}\frac{2\rvx^\gB}{1+\|\rvx^\gB\|^2}+\frac{\tanh(2\|\rvv^\gB\|)\frac{2\rvx^\gB\cdot2\rvv^\gB}{1+\|\rvx^\gB\|^2}}{2\|\rvv^\gB\|+\tanh(2\|\rvv^\gB\|)\frac{2\rvx^\gB\cdot2\rvv^\gB}{1+\|\rvx^\gB\|^2}}\rvx^\gB\\
&\qquad + \frac{\tanh(2\|\rvv^\gB\|)\sqrt{1-\frac{4\|\rvx^\gB\|^2}{\left(1+\|\rvx^\gB\|^2\right)^2}}}{2\|\rvv^\gB\|+\tanh(2\|\rvv^\gB\|)\frac{2\rvx^\gB\cdot2\rvv^\gB}{1+\|\rvx^\gB\|^2}}2\rvv^\gB\\
&= \frac{\|\rvv^\gB\|\left(1+\|\rvx^\gB\|^2\right)}{\|\rvv^\gB\|\left(1+\|\rvx^\gB\|^2\right)+2\tanh(2\|\rvv^\gB\|)\rvx^\gB\cdot\rvv^\gB}\frac{2\rvx^\gB}{1+\|\rvx^\gB\|^2}\\
&\qquad + \frac{2\tanh(2\|\rvv^\gB\|)\rvx^\gB\cdot\rvv^\gB}{\|\rvv^\gB\|\left(1+\|\rvx^\gB\|^2\right)+2\tanh(2\|\rvv^\gB\|)\rvx^\gB\cdot\rvv^\gB}\rvx^\gB\\
&\qquad + \frac{\tanh(2\|\rvv^\gB\|)\left(1-\|\rvx^\gB\|^2\right)}{\|\rvv^\gB\|\left(1+\|\rvx^\gB\|^2\right)+2\tanh(2\|\rvv^\gB\|)\rvx^\gB\cdot\rvv^\gB}\rvv^\gB\\
&= \frac{2\|\rvv^\gB\|+2\tanh(2\|\rvv^\gB\|)\rvx^\gB\cdot\rvv^\gB}{\|\rvv^\gB\|\left(1+\|\rvx^\gB\|^2\right)+2\tanh(2\|\rvv^\gB\|)\rvx^\gB\cdot\rvv^\gB}\rvx^\gB \\
&\qquad +\frac{\tanh(2\|\rvv^\gB\|)\left(1-\|\rvx^\gB\|^2\right)}{\|\rvv^\gB\|\left(1+\|\rvx^\gB\|^2\right)+2\tanh(2\|\rvv^\gB\|)\rvx^\gB\cdot\rvv^\gB}\rvv^\gB\\
&= \frac{2\|\rvv^\gB\|+\frac{4\tanh(\|\rvv^\gB\|)\rvx^\gB\cdot\rvv^\gB}{1+\tanh^2(\|\rvv^\gB\|)}}{\|\rvv^\gB\|\left(1+\|\rvx^\gB\|^2\right)+\frac{4\tanh(\|\rvv^\gB\|)\rvx^\gB\cdot\rvv^\gB}{1+\tanh^2(\|\rvv^\gB\|)}}\rvx^\gB+\frac{\frac{2\tanh(\|\rvv^\gB\|)\left(1-\|\rvx^\gB\|^2\right)}{1+\tanh^2(\|\rvv^\gB\|)}}{\|\rvv^\gB\|\left(1+\|\rvx^\gB\|^2\right)+\frac{4\tanh(\|\rvv^\gB\|)\rvx^\gB\cdot\rvv^\gB}{1+\tanh^2(\|\rvv^\gB\|)}}\rvv^\gB\\
&= \frac{2\|\rvv^\gB\|\left(1+\tanh^2(\|\rvv^\gB\|)\right)+4\tanh(\|\rvv^\gB\|)\rvx^\gB\cdot\rvv^\gB}{\|\rvv^\gB\|\left(1+\|\rvx^\gB\|^2\right)\left(1+\tanh^2(\|\rvv^\gB\|)\right)+4\tanh(\|\rvv^\gB\|)\rvx^\gB\cdot\rvv^\gB}\rvx^\gB\\
&\qquad +\frac{2\tanh(\|\rvv^\gB\|)\left(1-\|\rvx^\gB\|^2\right)}{\|\rvv^\gB\|\left(1+\|\rvx^\gB\|^2\right)\left(1+\tanh^2(\|\rvv^\gB\|)\right)+4\tanh(\|\rvv^\gB\|)\rvx^\gB\cdot\rvv^\gB}\rvv^\gB\\
&=  \frac{\left(2\|\rvv^\gB\|+2\tanh^2(\|\rvv^\gB\|)\|\rvv^\gB\|+4\tanh(\|\rvv^\gB\|)\rvx^\gB\cdot\rvv^\gB\right)\rvx^\gB+\left(2\tanh(\|\rvv^\gB\|-2\tanh(\|\rvv^\gB\|)\|\rvx^\gB\|^2\right)\rvv^\gB}{\|\rvv^\gB\|+\|\rvx^\gB\|^2\|\rvv^\gB\|+\tanh^2(\|\rvv^\gB\|)\|\rvv^\gB\|+\tanh^2(\|\rvv^\gB\|)\|\rvx^\gB\|^2\|\rvv^\gB\|+4\tanh(\|\rvv^\gB\|)\rvx^\gB\cdot\rvv^\gB}
\end{align*}

Meanwhile, we need to map $\rvx^{\gB}\oplus_{\text{M}}\exp^\gB_{\rvo^\gB}(\rvv^{\gB})\in \sB^n$ to $\sK^n$ and check whether it matches the simplified result of $\rvx^{\gK}\oplus_\text{E}\exp^\gK_{\rvo^\gK}(\rvv^{\gK})$.
Denote 
\begin{align}
\|\rvv^{\gB}\|+2\tanh{\left(\|\rvv^{\gB}\|\right)} \rvx^\gB\cdot \rvv^\gB+\tanh^2{\left(\|\rvv^{\gB}\|\right)}\|\rvx^\gB\|^2\|\rvv^{\gB}\| &= A,\\ \|\rvv^{\gB}\|+2\tanh{\left(\|\rvv^{\gB}\|\right)} \rvx^\gB\cdot \rvv^\gB+\tanh^2{\left(\|\rvv^{\gB}\|\right)}\|\rvv^{\gB}\| &= B,\\
\left(1-\|\rvx^\gB\|^2\right) \tanh{\left(\|\rvv^{\gB}\|\right)} &= C.
\end{align}
Then $\rvx^{\gB}\oplus_{\text{M}}\exp^\gB_{\rvo^\gB}(\rvv^{\gB})\in \sB^n$ can be written as $\frac{B\rvx^\gB+C\rvv^\gB}{A}$. Thus, the corresponding point in $\sK^n$ is given by 
\begin{equation}\label{eq:term of interest}
\begin{aligned}
\frac{2\frac{B\rvx^\gB+C\rvv^\gB}{A}}{1+\|\frac{B\rvx^\gB+C\rvv^\gB}{A}\|^2} &= \frac{2AB\rvx^\gB+2AC\rvv^\gB}{A^2+\|B\rvx^\gB+C\rvv^\gB\|^2}\\
&= \frac{2AB\rvx^\gB+2AC\rvv^\gB}{A^2+B^2\|\rvx^\gB\|^2+C^2\|\rvv^\gB\|^2+2BC(\rvx^\gB\cdot\rvv^\gB)}.
\end{aligned}
\end{equation}
For clarity, let $\|\rvx^\gB\|=x, \|\rvv^\gB\|=v$, and $\rvx^\gB\cdot\rvv^\gB=xv\cos{\theta}$. We calculate:
\begin{equation}
\begin{aligned}
A^2 &= (v+2xv\cos(\theta)\tanh(v)+x^2v\tanh^2(v))^2\\
&= v^2 + 4x v^2  \cos(\theta) \tanh(v) + 2 x^2 v^2 \tanh^2(v) \\
&\quad + 4 x^2v^2 \cos^2(\theta) \tanh^2(v) + 4 x^3 v^2 \cos(\theta) \tanh^3(v) +  x^4 v^2 \tanh^4(v),
\end{aligned}
\end{equation}
\begin{equation}
\begin{aligned}
B^2\|\rvx^\gB\|^2 &= \left(v + 2xv\cos(\theta)\tanh(v) + v\tanh^2(v)\right)^2x^2 \\
&= x^2 v^2 + 4 x^3 v^2 \cos(\theta) \tanh(v) + 2 x^2 v^2 \tanh^2(v) \\
&\quad + 4 x^4 v^2 \cos^2(\theta) \tanh^2(v) + 4 x^3 v^2 \cos(\theta) \tanh^3(v) + x^2 v^2 \tanh^4(v),
\end{aligned}    
\end{equation}
\begin{equation}
\begin{aligned}
C^2\|\rvv^\gB\|^2 = \left(1-x^2\right)^2v^2\tanh^2(v) = v^2 \tanh^2(v) - 2 x^2 v^2 \tanh^2(v) + x^4 v^2 \tanh^2(v),
\end{aligned}   
\end{equation}
\begin{equation}
\begin{aligned}
2BC(\rvx^\gB\cdot\rvv^\gB) &= 2xv\cos(\theta)\left(v + 2xv\cos(\theta)\tanh(v) + v\tanh^2(v)\right)\left(1-x^2\right)\tanh(v)\\
&= 2 x v^2 \cos(\theta) \tanh(v) - 2 x^3 v^2 \cos(\theta) \tanh(v) + 4 x^2 v^2 \cos^2(\theta) \tanh^2(v) \\
&\quad - 4 x^4 v^2 \cos^2(\theta) \tanh^2(v) + 2 x v^2 \cos(\theta) \tanh^3(v) - 2 x^3 v^2 \cos(\theta) \tanh^3(v).
\end{aligned}  
\end{equation}
Thus, the denominator in \equationref{eq:term of interest} is
\begin{equation}
\begin{aligned}
&\quad A^2+B^2\|\rvx^\gB\|^2+C^2\|\rvv^\gB\|^2+2BC(\rvx^\gB\cdot\rvv^\gB)\\
&= v^2 + x^2 v^2 + 6 x v^2 \cos(\theta) \tanh(v) + 2 x^3 v^2 \cos(\theta) \tanh(v) \\
&\quad + v^2 \tanh^2(v) + 2 x^2 v^2 \tanh^2(v) + x^4 v^2 \tanh^2(v) + 8 x^2 v^2 \cos^2(\theta) \tanh^2(v) \\
&\quad + 2 x v^2 \cos(\theta) \tanh^3(v) + 6 x^3 v^2 \cos(\theta) \tanh^3(v) + x^2 v^2 \tanh^4(v) + x^4 v^2 \tanh^4(v).
\end{aligned} 
\end{equation}
We can factor this expression and get 
\begin{equation}
\begin{aligned}
&\quad A^2+B^2\|\rvx^\gB\|^2+C^2\|\rvv^\gB\|^2+2BC(\rvx^\gB\cdot\rvv^\gB)\\
&= v^2 \left(1 + 2 x \cos(\theta) \tanh(v) + x^2 \tanh^2(v)\right) \\
&\qquad \cdot \left(1 + x^2 + 4 x \cos(\theta) \tanh(v) + \tanh^2(v) + x^2 \tanh^2(v)\right).
\end{aligned}
\end{equation}
The numerator contains the following terms:
\begin{align}
2AB &= 2\left(v + 2 xv \cos(\theta) \tanh(v) + x^2v \tanh^2(v)\right)\left(v + 2xv\cos(\theta)\tanh(v) + v\tanh^2(v)\right) \nonumber \\
&= 2v^2\left(1 + 2 x \cos(\theta) \tanh(v) + x^2\tanh^2(v)\right)\left(1 + 2x\cos(\theta)\tanh(v) + \tanh^2(v)\right),\\
2AC &= 2\left(v + 2 xv \cos(\theta) \tanh(v) + x^2v \tanh^2(v)\right)\left(1-x^2\right)\tanh(v) \nonumber \\
&= 2v\left(1 + 2 x \cos(\theta) \tanh(v) + x^2 \tanh^2(v)\right)\left(\tanh(v)-x^2\tanh(v)\right).
\end{align}
Altogether, \equationref{eq:term of interest} can be simplified as 
\begin{align}
&\quad \frac{2AB\rvx^\gB+2AC\rvv^\gB}{A^2+B^2\|\rvx^\gB\|^2+C^2\|\rvv^\gB\|^2+2BC(\rvx^\gB\cdot\rvv^\gB)} \nonumber \\
&= \frac{2v^2\left(1 + 2 x \cos(\theta) \tanh(v) + x^2\tanh^2(v)\right)\left(1 + 2x\cos(\theta)\tanh(v) + \tanh^2(v)\right)}{v^2 \left(1 + 2 x \cos(\theta) \tanh(v) + x^2 \tanh^2(v)\right) \left(1 + x^2 + 4 x \cos(\theta) \tanh(v) + \tanh^2(v) + x^2 \tanh^2(v)\right)}\rvx^\gB \nonumber \\
&\quad + \frac{2v\left(1 + 2 x \cos(\theta) \tanh(v) + x^2 \tanh^2(v)\right)\left(\tanh(v)-x^2\tanh(v)\right)}{v^2 \left(1 + 2 x \cos(\theta) \tanh(v) + x^2 \tanh^2(v)\right) \left(1 + x^2 + 4 x \cos(\theta) \tanh(v) + \tanh^2(v) + x^2 \tanh^2(v)\right)}\rvv^\gB \nonumber \\
&= \frac{2\left(1 + 2x\cos(\theta)\tanh(v) + \tanh^2(v)\right)}{1 + x^2 + 4 x \cos(\theta) \tanh(v) + \tanh^2(v) + x^2 \tanh^2(v)}\rvx^\gB  \nonumber \\
&\quad +\frac{2\left(\tanh(v)-x^2\tanh(v)\right)}{v\left(1 + x^2 + 4 x \cos(\theta) \tanh(v) + \tanh^2(v) + x^2 \tanh^2(v)\right)}\rvv^\gB. \label{eq:compare1}
\end{align} 
If we express $\rvx^{\gK}\oplus_\text{E}\exp^\gK_{\rvo^\gK}(\rvv^{\gK})$ in terms of $x, v$ and $\cos(\theta)$, we get
\begin{align}
& \quad \rvx^{\gK}\oplus_\text{E}\exp^\gK_{\rvo^\gK}(\rvv^{\gK}) \nonumber \\ 
&= \frac{\left(2\|\rvv^\gB\|+2\tanh^2(\|\rvv^\gB\|)\|\rvv^\gB\|+4\tanh(\|\rvv^\gB\|)\rvx^\gB\cdot\rvv^\gB\right)\rvx^\gB+\left(2\tanh(\|\rvv^\gB\|-2\tanh(\|\rvv^\gB\|)\|\rvx^\gB\|^2\right)\rvv^\gB}{\|\rvv^\gB\|+\|\rvx^\gB\|^2\|\rvv^\gB\|+\tanh^2(\|\rvv^\gB\|)\|\rvv^\gB\|+\tanh^2(\|\rvv^\gB\|)\|\rvx^\gB\|^2\|\rvv^\gB\|+4\tanh(\|\rvv^\gB\|)\rvx^\gB\cdot\rvv^\gB} \nonumber \\ 
&= \frac{2\left(v + 2xv\cos(\theta)\tanh(v) + v\tanh^2(v)\right)\rvx^\gB+ 2\left(\tanh(v)-x^2\tanh(v)\right)\rvv^\gB}{v + x^2v + 4 xv \cos(\theta) \tanh(v) + \tanh^2(v)v + x^2v \tanh^2(v)} \nonumber \\ 
&= \frac{2\left(1 + 2x\cos(\theta)\tanh(v) + \tanh^2(v)\right)}{1 + x^2 + 4 x \cos(\theta) \tanh(v) + \tanh^2(v) + x^2 \tanh^2(v)}\rvx^\gB  \nonumber \\ 
&\quad +\frac{2\left(\tanh(v)-x^2\tanh(v)\right)}{v\left(1 + x^2 + 4 x \cos(\theta) \tanh(v) + \tanh^2(v) + x^2 \tanh^2(v)\right)}\rvv^\gB. \label{eq:compare2}
\end{align}

The expressions in \equationref{eq:compare1} and \equationref{eq:compare2} agree. Therefore, $\rvx^{\gK}\oplus_{\text{E}}\exp^\gK_{\rvo^\gK}(\rvb^{\gK})$ and $\rvx^{\gB}\oplus_{\text{M}}\exp^\gB_{\rvo^\gB}(\rvb^{\gB})$ are the corresponding points in the hyperbolic space. We conclude that 
\begin{equation}
P_{\rvo^{\gK}\to\rvx^{\gK}}(\rvv^{\gK}) = \log^\gK_{\rvx^\gK}\left(\rvx^{\gK}\oplus_\text{E}\exp^\gK_{\rvo^\gK}(\rvv^{\gK})\right).
\end{equation}
\qed

\subsection{Proof of \theoremref{thm:Einstein matrix-vector multiplication}}
A simple application of \equationref{eq:exp_log_K_0} yields 
\begin{equation}
\begin{aligned}
\rmM^{\otimes_\text{E}}(\rvx^\gK) &= \exp^\gK_{\rvo^\gK}(\rmM\log^\gK_{\rvo^\gK}( \rvx^\gK)) \\
&= \tanh \left(\frac{2\|\rmM \rvx^\gK\|}{\|\rvx^\gK\|} \tanh ^{-1}\left(\frac{\|\rvx^\gK\|}{1+\sqrt{1-\|\rvx^\gK\|^2}}\right)\right) \frac{\rmM \rvx^\gK}{\|\rmM\rvx^\gK\|}.
\end{aligned}
\end{equation}
\qed

\newpage
\section{Statistics of Datasets}\label{apd:third}

The following table summarizes the statistics of the datasets.
\begin{table}[hbtp]
\floatconts
  {tab:statistics}
  \centering
  {\caption{Statistics of datasets}\label{tab:statistics}}
  \footnotesize
  {\begin{tabular}{l|llllll}
  \toprule
  \bfseries Dataset & \bfseries Texas & \bfseries Wisconsin & \bfseries Chameleon & \bfseries Actor & \bfseries Cora  &\bfseries Pubmed\\
  \midrule
  Nodes     & 183   & 251   & 2,277  & 7,600  & 2,708 & 19,717   \\
Edges & 280   & 466   & 31,421   & 26,752 & 5,278 & 44,327\\
Features & 1,703   & 1,703   & 2,325 & 931 & 1,433  &  500  \\
Class & 5 & 5 & 5 & 5 & 7 & 3 \\
Hyperbolicity & 1.0  & 1.0 & 1.5  & 1.5 & 3.0 & 2.5\\
  \bottomrule
  \end{tabular}}
\end{table}

\end{document}